\documentclass[twoside,leqno,twocolumn]{article}

\usepackage[letterpaper]{geometry}

\usepackage{ltexpprt}
\usepackage{hyperref}

\usepackage{amsmath,amssymb,amsfonts}
\usepackage[utf8]{inputenc}
\usepackage[english]{babel}
\usepackage{subfigure}
\usepackage{mathtools}
\usepackage[algo2e, ruled, linesnumbered]{algorithm2e}
\usepackage{bm}
\usepackage{booktabs}
\usepackage{balance}
\usepackage{makecell}
\usepackage{paralist}

\newcommand{\method}{\textit{AutoNR}}

\begin{document}
\newcommand\relatedversion{}

\title{\Large Automatic Parameter Selection for Non-Redundant Clustering\relatedversion}
\author{Collin Leiber\thanks{LMU Munich. \{leiber, mautz, boehm\}@dbs.ifi.lmu.de}
	\and Dominik Mautz$^\ast$
	\and Claudia Plant\thanks{Faculty of Computer Science, ds:UniVie, University of Vienna, Vienna, Austria. claudia.plant@univie.ac.at}
	\and Christian Böhm$^{\ast}$}

\date{}

\maketitle

\fancyfoot[R]{\scriptsize{Copyright \textcopyright\ 2022 by SIAM\\
		Unauthorized reproduction of this article is prohibited}}

\begin{abstract} \small\baselineskip=9pt 
High-dimensional datasets often contain multiple meaningful clusterings in different subspaces. For example, objects can be clustered either by color, weight, or size, revealing different interpretations of the given dataset.
A variety of approaches are able to identify such non-redundant clusterings. However, most of these methods require the user to specify the expected number of subspaces and clusters for each subspace. Stating these values is a non-trivial problem and usually requires detailed knowledge of the input dataset. In this paper, we propose a framework that utilizes the Minimum Description Length Principle (MDL) to detect the number of subspaces and clusters per subspace automatically. 
We describe an efficient procedure that greedily searches the parameter space by splitting and merging subspaces and clusters within subspaces.
Additionally, an encoding strategy is introduced that allows us to detect outliers in each subspace. Extensive experiments show that our approach is highly competitive to state-of-the-art methods.
\end{abstract}

\section{Introduction}

\begin{figure}[t]
	\centering
	\includegraphics[width=0.34\textwidth]{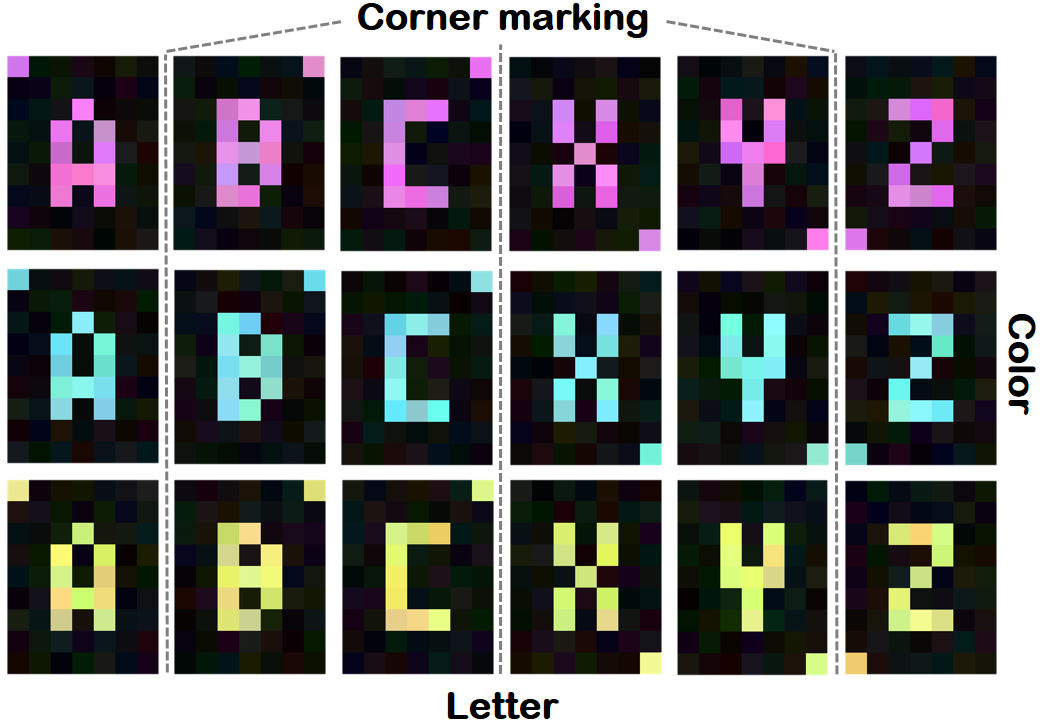}
		\caption{Images of the letters 'A', 'B', 'C', 'X', 'Y' and 'Z' in the colors pink, cyan, and yellow. In each image, one corner is highlighted in color. This results in three different clusterings with 6, 3, and 4 clusters.}
	\label{fig:nrLetters}
\end{figure}

Several algorithms have been developed in order to cluster objects under certain conditions. Typically, these techniques deliver a single solution or multiple solutions in a hierarchical setup. However, often various distinct clusterings can be created by different characteristics of the given dataset. Take Figure \ref{fig:nrLetters} as an example. These low-resolution images can be grouped by the shown letter, the color, or the marked corner. This indicates that there are three different lower-dimensional subspaces, each with a unique clustering. Further, these partitionings are mutually non-redundant, for each object belongs to different clusters in different subspaces \cite{mautz2018discovering}.

This diversity of clustering possibilities emerges especially from datasets in high-dimensional spaces. Non-redundant clustering algorithms can reveal coherences within the data, which would not be visible by a single clustering. Since each clustering has its individual subspace, the analysis of the results is facilitated by the lower dimensionality and usually also by to the lower number of clusters. In addition, the user can choose the best partitioning based on his specific objective without being limited to a single option. However, a problem with most available methods is that the user needs a lot of prior knowledge to set suitable input parameters. 

In this paper, we describe a framework that utilizes the Minimum Description Length (MDL) to overcome the problem of specifying those parameters. MDL is used to solve model selection problems and for example is often applied in Boolean Matrix Factorization \cite{miettinen2011model}. It is based on the idea that any regularity in the data can be used to compress the data \cite{grunwald2005tutorial}. The fewer bits are required to describe a model, the better the model is. 

Our framework performs a sophisticated search in the parameter space to find a setting with which a high compression of the input dataset is achieved. Thus, it can automatically determine the number of subspaces and clusters per subspace. Existing parameter-free approaches (e.g. \cite{wang2019multiple, ye2016generalized}) first identify all subspaces and then the clusters within them. In contrast, we successively split and merge subspaces while simultaneously searching for the best number of partitions within these new subspaces. This gives us more flexibility in finding meaningful structures. Additionally, we introduce a feature that can identify outliers in each of the found subspaces. This feature is integrated into our MDL-based cost function and does not require additional input parameters. To illustrate the effectiveness of our framework we created the \method~algorithm which combines our proposal with the recently presented \textit{Nr-Kmeans} \cite{mautz2018discovering} approach. Our major contributions are:
\begin{itemize}
	\item We present a general MDL encoding for non-redundant clustering models that can be combined with centroid-based clustering algorithms.
	\item A parameter-free, greedy approach efficiently searches the parameter space to identify the number of subspaces and clusters within subspaces. 
	\item Our procedure is extended to identify outliers in each existing subspace.
\end{itemize}

\section{Related Work}
\label{chap:relatedWork}

We want to briefly review three relevant areas of research: non-redundant clustering, methods for determining the number of clusters and basic outlier detection extensions for Gaussian clustering approaches.

\noindent \textbf{Non-Redundant Clustering:} We only consider methods that can find non-redundant clusterings without knowing an initial clustering solution. Alternative clustering algorithms would require an additional parameter-free clustering approach that provides the input partitioning in a parameter-free setting.

Cui et al. \cite{cui2007non} identify orthogonal clusterings by sequentially executing PCA and \textit{k-means}. They propose two strategies: \textit{orthogonal clustering (Orth1)} and \textit{clustering in orthogonal subspaces (Orth2)}. \textit{mSC} \cite{niu2010multiple} creates non-redundant partitionings by combining spectral clustering with the Hilbert-Schmidt Independence Criterion to penalize similarity between different subspaces. \textit{Nr-Kmeans} \cite{mautz2018discovering}, a generalization of the subspace clustering algorithm \textit{SubKmeans} \cite{mautz2017towards}, splits a rotated feature space into $J$ axis-parallel subspaces, each containing an individual \textit{k-means} clustering.
An advantage is that it optimizes all subspaces and clusters in the subspaces simultaneously. 
\textit{NrDipmeans} \cite{mautz2020non} is an extension of this method. Here, Hartigan's dip test \cite{hartigan1985dip} is used to detect multimodal structures in each subspace and thereby determine the number of clusters. The five algorithms mentioned so far all require the number of subspaces and, except for \textit{NrDipmeans}, the number of clusters per subspace as input parameters.
\textit{ISAAC} \cite{ye2016generalized} utilizes Independent Subspace Analysis (ISA) and MDL to find multiple subspaces in a parameter-free setting. It fits Gaussian Mixture Models (GMMs) within those statistically independent subspaces. The parameter-free algorithm \textit{MISC} \cite{wang2019multiple} also uses ISA and MDL to identify independent subspaces. Afterward, it performs kernel graph regularized semi-nonnegative matrix factorization to define clusters in each subspace.

\noindent \textbf{Determine the Number of Clusters:} A popular strategy to identify the number of clusters is the \textit{X-means} algorithm \cite{pelleg2000x}. It uses the Bayesian Information Criterion to rate different parametrizations of \textit{k-means}. Starting with a small number of clusters, it repeatedly splits them up by creating two new centers along a randomly chosen vector. Ishioka et al. \cite{ishioka2005expansion} extend this approach by introducing the idea of merging clusters.
The \textit{G-means} \cite{hamerly2004learning} algorithm exploits the Anderson-Darling statistical hypothesis test to check whether the objects within a cluster follow a Gaussian distribution. The approach starts with a single cluster and iteratively increases the number of clusters by splitting those whose data appears non-Gaussian.
\textit{PG-means} \cite{feng2007pg} projects the data and the learned model into a one-dimensional space and applies the Kolmogorov-Smirnov test to check if the projected model fits the projected data. This test is executed repeatedly with different projections. If any test fails, a new cluster will be added, and the approach repeats.
Bischof et al. \cite{bischof1999mdl} use MDL as a pruning criterion by starting with a complex model and gradually reducing the number of clusters until optimal encoding costs are reached. Additionally, they use MDL to identify outliers in each iteration of their procedure. 
\textit{FOSSCLU} \cite{goebl2014finding} is a subspace clustering algorithm optimizing an orthonormal transformation that projects the data into a lower-dimensional subspace in which the \textit{EM algorithm} is executed. The dimensionality of the subspace and the number of clusters are identified through MDL.

Most aboved-mentioned methods cannot be applied easily to non-redundant clustering, since the subspaces are constantly adapting to changing cluster structures. This particularly affects the search heuristics.

\noindent \textbf{Outlier Detection:} The identification of outliers usually requires additional input parameters. For example, \textit{ODIN} \cite{hautamaki2004outlier} first creates a k-nearest neighbor graph and then counts the number of edges leading to a point. If this amount is smaller than a threshold, the point is an outlier.
\textit{ORC} \cite{hautamaki2005improving} calculates the outlyingness of all points as their distance to their corresponding cluster centers divided by the maximum distance. If the outlyingness of a point is greater than a threshold, it will be removed from the dataset.
\textit{k-means-{}-} \cite{chawla2013k} regards the $l$ points farthest away from their corresponding centers as outliers in each iteration of \textit{k-means}. These $l$ points will not be used for the subsequent assignment and update step.
\textit{ODC} \cite{ahmed2013novel} and \textit{KMOR} \cite{gan2017k} both calculate the average distance $d_{avg}$ between the objects and the cluster centers. If an object is at least $y \cdot d_{avg}$ away from all centers, with $y$ being a parameter set by the user, it will be regarded as an outlier. One difference is that \textit{KMOR} allows outliers to become inliers again.

For our parameter-free approach, it is important that no additional parameter for the outlier identification is introduced. Therefore, we integrate the outlier definition directly into our MDL-based cost function.

\section{Parameter-free Non-Redundant Clustering}
\label{chap:paramterFreeNrKmeans}

While algorithms like \textit{Orth} \cite{cui2007non}, \textit{mSC} \cite{niu2010multiple}, and \textit{Nr-Kmeans} \cite{mautz2018discovering} work well in defining non-redundant clusterings, they still require extensive user knowledge of the input dataset to define the correct number of subspaces $J$ and clusters per subspace $k_j$.

In the following we want to use MDL to solve this situation. All
symbols used in this work are summarized and described in the supplement. 

Usually, the subspace of a non-redundant clustering algorithm can be defined using an orthogonal transformation matrix $V \in \mathbb{R}^{d\times d}$ that rotates the given $d$-dimensional feature space and a projection matrix $P_j \in \mathbb{N}^{d\times m_j}$ that specifies its corresponding $m_j$ dimensions after rotation. $P_j$ has a $1$ in its $i$-th row if the $i$-th dimension is contained in subspace $j$. All other values are $0$. In addition, the sum of each column must be equal to $1$.
Consequently, in subspace $j$ the input dataset $X \subseteq \mathbb{R}^{d}$ of size $N$ changes to
\begin{equation}
X_j = \{xVP_j | x \in X\},
\label{eq:newX}
\end{equation}
where $X_j \subseteq \mathbb{R}^{m_j}$.
The actual clustering then takes place in these subspaces.
This is a common setting for many subspace and non-redundant clustering algorithms, like those presented in \cite{cui2007non,goebl2014finding,mautz2017towards,mautz2018discovering,niu2010multiple,wang2019multiple,ye2016generalized}. 

\subsection{MDL Encoding}
\label{sec:mdlModelCosts}

First, we describe our MDL encoding strategy for non-redundant clusterings.
In general, MDL tries to minimize $L(H)+L(D|H)$. $L(H)$ indicates the number of bits necessary to define the hypothesis, which in our case equals the rotation, subspace, and cluster parameters. $L(D|H)$ is the code length needed to encode the data under this hypothesis \cite{grunwald2005tutorial}. 

We can encode the natural numbers $J$, $m_j$, and $k_j$ with the universal prior for integers, $L^0(n)$ \cite{rissanen1983universal}.
\begin{equation*}
L^0(n) = \log^*(n) + \log_2(c),
\end{equation*}
where $\log^*(n) = \log_2(n) + \log^*(\log_2(n))$ only involves the positive terms and $c \simeq 2.865064$. 

The cluster centers in subspace $j$ are generally represented by a vector consisting of $m_j$ real values, each corresponding to one coordinate of the rotated feature space. To encode these values, we utilize a uniform distribution and apply Shannon-Fano coding \cite{shannon1948mathematical}. By using the uniform distribution, no position in the feature space is favored. To implement this, we first calculate the maximum distance between all objects. In combination with the minimum coordinate of each rotated feature, we can form a hypercube that contains all points with certainty, regardless of the final rotation. The resulting probability density function (pdf) looks as follows:
\begin{equation*}
\pi_{\text{uniform}}(r)=\frac{1}{\text{max}(\text{dist}_X)},
\end{equation*}
where $\text{dist}_X$ is the set of Euclidean distances between all objects, $\text{dist}_X = \{\sqrt{||x-y||_2^2} ~ | ~ x \in X \text{ and } y \in X\}$.

In order to correctly apply Shannon-Fano codes, we need to transform the result of a pdf to an actual probability. This can approximately be done by using a precision $\delta$ \cite{lee2001introduction}. We assume that each object origins from a creation process that itself operates with a finite precision. Note, that the creation process of each feature can be different; therefore the precision can vary greatly. Since the multiplication by $V$ creates a new space in which all features are mixed to some degree, we approximate $\delta$ by using the mean of the minimum feature-wise distances (ignoring zero values). 
The total code length is a combination of the pdf and $\delta$ \cite{lee2001introduction}.
\begin{equation*}
L(r)=-\log_2(\pi(r))-\log_2(\delta)
\end{equation*}
We use this to get the coding costs of cluster center $\mu_{j, i}$ in subspace $j$.
\begin{equation}
\label{eq:clusterCenter}
L(\mu_{j, i})=-m_j\left(\log_2(\pi_{\text{uniform}}(\mu_{j,i}))+\log_2(\delta)\right)
\end{equation}

To encode the cluster assignments, we assume that the probability of belonging to a particular cluster in subspace $j$ is the same for all clusters, and therefore equals $\frac{1}{k_j}$. This way, we can again utilize Shannon–Fano coding and receive a code length of $-\log_2(\frac{1}{k_j})N$ \cite{shannon1948mathematical}.

Note that the costs of encoding $N$, $d$, $V$, and the features of the hypercube are always the same, regardless of which model we are analyzing. Therefore, these constant costs can be ignored.

The objects in a dataset can be encoded by choosing a suitable distribution function matching the clustering procedure. This could, for example, be a Gaussian Mixture Model (GMM) for \textit{EM}-like algorithms. It is important that the distribution fits the underlying clustering algorithm, so that our parameter-search procedure does not work against that algorithm by trying to optimize something different. Once a suitable pdf ($\pi_\text{obj}$) has been identified, the coding costs of the objects in subspace $j$ are as follows:
\begin{equation}
\begin{gathered}
\label{eq:encodingCosts}
L(X_j) = \sum_{x \in X_j}\left(-\log_2(\pi_\text{obj}(x))-m_j\log_2(\delta)\right)
\end{gathered}
\end{equation}

Finally, the distribution-specific parameters have to be encoded. For a GMM, for example, these would be the $k_jd(d+1)/2$ values for the symmetric covariance matrices $\Sigma_{j,i}$ of the clusters in subspace $j$. Let's assume we have $p_j$ such parameters in subspace $j$. Since we normally use a maximum likelihood estimator to calculate those values, the code length can be approximated by $\frac{p_j}{2}\log_2(N)$ \cite{rissanen1986stochastic, lee2001introduction}. We use $N$ because it is a well-known value that usually is sufficiently large.

In summary, the costs required to encode a complete non-redundant clustering model in bits are:
\begin{compactitem}
	\item \makebox[4cm][l]{number of subspaces:} $L^0(J)$
	\item for each subspace $j$:
	\begin{compactitem}
		\item \makebox[3.8cm][l]{dimensionality:} $L^0(m_j)$
		\item \makebox[3.8cm][l]{number of clusters:} $L^0(k_j)$
		\item \makebox[3.8cm][l]{cluster centers:} \makebox[2.2cm][l]{$\sum_{i=1}^{k_j} L(\mu_{j,i})$} (\ref{eq:clusterCenter})
		\item \makebox[3.8cm][l]{cluster assignments:} $-\log_2(\frac{1}{k_j})N$
		\item \makebox[3.8cm][l]{objects:} \makebox[2.2cm][l]{$L(X_j)$} (\ref{eq:encodingCosts})
		\item \makebox[3.8cm][l]{distribution parameters:} $\frac{p_j}{2}\log_2(N)$
	\end{compactitem}
\end{compactitem}

\subsection{Determine the Subspace and Cluster Count}

We use this encoding strategy to develop a greedy procedure that identifies the number of subspaces and clusters within each subspace in a non-redundant clustering setting. For this, multiple models have to be computed and evaluated by their MDL costs. Theoretically, the number of subspaces can range from $1$ to $d$, and the number of clusters within each subspace from $1$ to $N$. This results in an extremely large search space. 

Our proposed strategy repeatedly splits the feature space into smaller subspaces until the MDL costs cannot be lowered anymore. Within those subspaces, we search for the best amount of clusters. Since subspaces are examined separately, the execution time drops significantly due to the lower dimensionality of the analyzed space. We differentiate between an optional subspace with a single cluster, called \textit{noise space}, and subspaces with more than one cluster, called \textit{cluster spaces}. The pseudo-code of the process is shown in the supplement.

We start with a random $V$ and a single \textit{noise space}.
Each iteration begins by sorting the \textit{cluster spaces} from the best full-space result so far in descending order by their MDL costs. An optional \textit{noise space} is added last. The ordering is used because costly subspaces offer the greatest potential for cost reduction through successive splitting operations. Only the corresponding subspace $j$ is used for these operations. This is done by changing the input from $X$ to the lower-dimensional $X_j$ as defined in Eq. (\ref{eq:newX}).
Depending on the type of subspace, we execute a \textit{noise} or \textit{cluster space} split.

\subsubsection{Noise Space Split}

In case of a \textit{noise space} split, the first run is performed with a subspace containing two clusters and a new lower-dimensional \textit{noise space}. In the following iterations, the number of clusters within the \textit{cluster space} is constantly raised by one. Here, the projections $P_j$ and the rotation matrix $V$ from the last result are used as input for the next iteration. We also take the cluster with the largest dispersion and divide it into two by creating two new centers. For cluster $i$ in subspace $j$, they are calculated as follows:
\begin{equation*}
\mu _{j, \text{new}_{1\text{ / }2}} = \mu_{j, i} \pm \frac{\text{diag}(\Sigma_{j, i})}{m_j|C_{j, i}|},
\end{equation*}
where $|C_{j, i}|$ is the size of the cluster and $\Sigma_{j, i}$ is its covariance matrix.
This allows us to use the centers as input too.

In addition to the execution with reused parameters, random parameterizations are applied to identify previously not found structures in the \textit{noise space}. In the following, we consider only the outcome that produces the lowest MDL cost.
If the newly created \textit{noise space} does not change in two consecutive iterations, we can stop using random parameters. In this case, we assume that a good structure in the \textit{cluster space} has been identified, but the number of clusters is yet uncertain. The \textit{noise space} splitting procedure repeats until the MDL costs increase.

\subsubsection{Cluster Space Split}

For \textit{cluster spaces}, we restrict the search space of possible splits. We assume that no arising subspace may have more clusters than the original space and that there cannot be more clusters in the original space than the number of possible combinations of clusters in the two resulting spaces. This gives the following rules:
\begin{equation}
	\text{max}(k_{\text{split}_1} , k_{\text{split}_2}) \leq k_{\text{original}} \le k_{\text{split}_1} \cdot k_{\text{split}_2}
	\label{eq:cluster_space_split}
\end{equation}

In the beginning, both subspaces have a cluster count equal to the original space. Subsequently, we merge the two nearest cluster centers in both subspaces by replacing them with their mean. Additionally, the projections $P_j$ and rotation matrix $V$ are reused as inputs for the following iteration. This approach repeats until Eq. (\ref{eq:cluster_space_split}) is violated. The idea is that the MDL costs reach a local minimum if the subspace with the higher number of clusters is correctly identified. Hereafter, we fix $k_j$ of the subspace that is more responsible for the low costs and successively reduce the number of clusters of the other subspace. This is repeated until Eq. (\ref{eq:cluster_space_split}) is violated, or the MDL costs increase. In the end, we output the result that produced the lowest overall costs.

\subsubsection{Full-space Execution}

When the subspace splitting procedure converges, we check whether the sum of the costs of the two new subspaces is smaller than the costs of the original space. If this is not the case, we start to split the next of the sorted subspaces or, if there is no subspace left, we start the merging operation. This cost-control mechanism avoids unnecessary expensive clustering executions in the full-space. After a positive check, a full-space clustering execution, which reuses the parameters of the last full-space result, is performed. Here, the parameters of the original subspace are replaced with the ones from the two subspaces found by the splitting procedure. The cluster centers can be brought to full dimensionality by using the cluster labels to calculate the mean of the assigned objects. The rotation matrix of the new subspaces can be converted into the full-space by using the following transformation formula \cite{mautz2018discovering}:
\begin{equation*}
\label{eq:toFull}
V_{\text{sub}}^{F}[a,b] = \begin{cases}
V_{\text{sub}}[n,m],& \text{if } P_{\text{sub}} \text{ maps $a$ to $n$ and $b$ to $m$}\\
1, &\text{if $a$ = $b$ and not the first case}\\
0, &\text{otherwise}
\end{cases}
\end{equation*}
We update the rotation matrix $V$ by calculating $V_\text{new} = V_\text{old}V_\text{sub}^F$. A single full-space execution is sufficient because all input parameters are known.

Some non-redundant clustering methods are not able to optimize all subspaces simultaneously. In this case, the full-space clustering execution can be skipped, and the model costs can be calculated directly using the combined parameters.

If the full-space solution has lower MDL costs than the best result found so far, a new iteration starts with the ordering of the subspaces. Otherwise, the result is discarded, and the next subspace is considered for splitting, or if no subspace is left, the merging process starts.

\subsubsection{Cluster Space Merge}

The merging step is executed for each possible combination of \textit{cluster spaces}. Again, the input dataset is transformed using Eq. (\ref{eq:newX}) with the combined projection matrices. Since we analyze the merged space individually, no projection or rotation matrices are needed. In the first iteration, we divide the dataset into $k_{\text{original}_1}\cdot k_{\text{original}_2}$ clusters. We use all combinations of centers of the two subspaces as the input set of centers. In each iteration, the number of clusters decreases by one. For this purpose, the nearest centers are merged by replacing them with their mean. The procedure stops if either the MDL costs rise or the cluster count gets smaller than the maximum number of clusters of the original subspaces. This equals the reversed rules of the \textit{cluster space} split from Eq. (\ref{eq:cluster_space_split}).
If the MDL costs of the merged space are lower than the sum of the costs of the original subspaces, we perform a full-space execution as described above.

If merging led to a better result, the merging operation repeats with the newly defined subspaces. Otherwise, our approach terminates. 

\begin{figure}[t]
	\centering
	\includegraphics[width=0.44\textwidth]{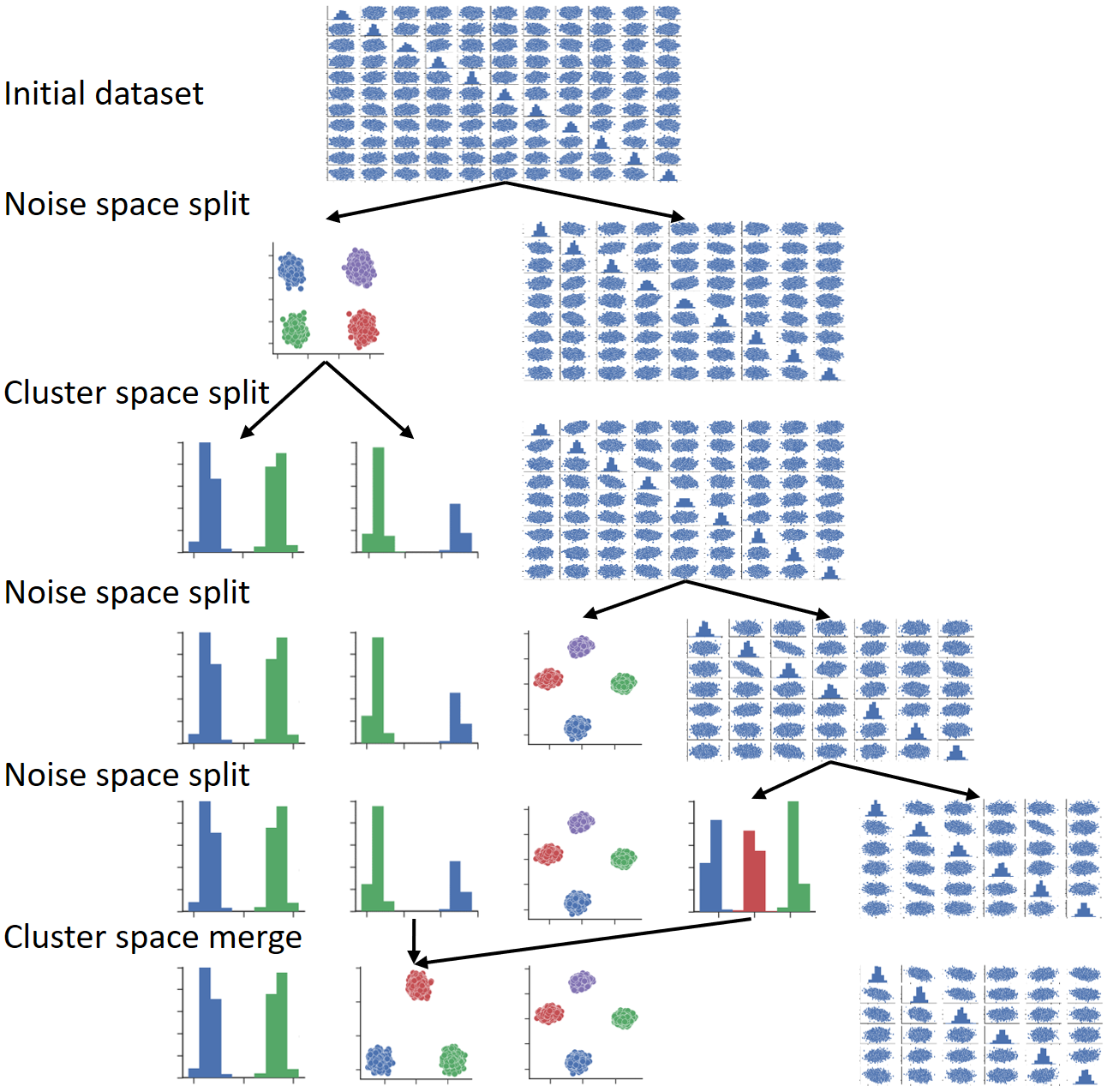}
	\caption{Example execution of \method~on a sample dataset ($d=11$). The arrows indicate which subspaces are affected by an operation. After two \textit{noise space} splits, a \textit{cluster space} split and a \textit{cluster space} merge, three \textit{cluster spaces} are identified with four, three, and two clusters.}
	\label{fig:example}
\end{figure}

An illustration of the described procedure is shown in Figure \ref{fig:example}. It visualizes the full-space results of our exemplary algorithm \method~(see Section \ref{chap:algorithm}) on a sample dataset. \method~first runs a \textit{noise space} split on the input dataset and finds a subspace with four clusters. Next, it performs a \textit{cluster space} split and splits the subspace into two subspaces with two clusters each. Then another \textit{noise space} split is performed, and a \textit{cluster space} with four clusters is found. The following \textit{noise space} split provides a subspace with three clusters. Last, this is merged with one of the subspaces with two clusters, resulting in three \textit{cluster spaces} with four, three, and two clusters.

\subsection{Outlier Detection}
Many clustering approaches suffer from outliers. For this reason, we introduce an optional outlier detection by using the ability to rate clustering results by their MDL costs. In the case of non-redundant clusterings, it is crucial to re-determine the outliers with each update of the cluster structure because the definition can change significantly due to an update of the rotation matrix or the projections. 

The cluster assignment and center update steps can be executed in the usual way. Afterward, we check for each point if it is cheaper to encode it using the distribution function of the corresponding cluster or separately as an outlier.
The following must hold for each outlier $o_j$ in subspace $j$:
\begin{equation*}
L(X_j \setminus o_j) + L_\text{total}(o_j) < L(X_j).
\end{equation*}

We can encode a single outlier $o_j$ in subspace $j$ the same way we encode the cluster centers in Section \ref{sec:mdlModelCosts}.
\begin{equation}
\label{eq:outlier}
L(o_j)=-m_j\left(\log_2(\pi_\text{uniform}(o_j))+\log_2(\delta)\right)
\end{equation}

In addition, we must consider that although we save one cluster assignment (requiring $-\log_2(\frac{1}{k_j})$ bits), we must specify the index of the specific outlier. This requires $\log_2(N)$ bits. 

Alternatively, one could also specify for each point whether it is an outlier or not (requiring $N$ bits) or interpret the outliers as an additional cluster and therefore encode the cluster assignments using $k_j+1$ groups. However, since it can be assumed that there are significantly fewer outliers than inliers, this would greatly increase the MDL costs.

Overall, the following costs decide whether or not it is worth interpreting a point as an outlier:
\begin{equation*}
L_\text{total}(o_j) = L(o_j) +\log_2(N) + \log_2(\frac{1}{k_j}).
\end{equation*}

After we identified all the outliers $O_j$ in subspace $j$, we update the cluster components (e.g. centers and covariance matrices).
Furthermore, we must slightly adapt the MDL costs of our clustering model. For each subspace $j$, we need to state the number of outliers, requiring $L^0(|O_j|)$ bits. Next, we must use $|O_j|\log_2(N)$ bits to encode the indices, and Eq. (\ref{eq:outlier}) has to be added $|O_j|$ times to encode the actual points. In return, only $N-|O_j|$ points have to be regarded for the cluster assignments.
The differences compared to the encoding summarized in Section \ref{sec:mdlModelCosts} are for each subspace:

\begin{itemize}
		\item[--] \makebox[3.4cm][l]{number of outliers:} $L^0(|O_j|)$
		\item[--] \makebox[3.4cm][l]{cluster assignments:} $-\log_2(\frac{1}{k_j})(N-|O_j|)$
		\item[--] \makebox[3.4cm][l]{indices of the outliers:} $|O_j|\log_2(N)$
		\item[--] \makebox[3.4cm][l]{outliers:} \makebox[2.5cm][l]{$\sum_{o_j\in O_j}L(o_j)$} (\ref{eq:outlier})
		\item[--] \makebox[3.4cm][l]{objects:} \makebox[2.5cm][l]{$L(X_j\setminus O_j)$} (\ref{eq:encodingCosts})
\end{itemize}

\begin{figure}[t]
	\centering
	\subfigure[Result without outlier detection.]{
		\includegraphics[width=0.22\textwidth]{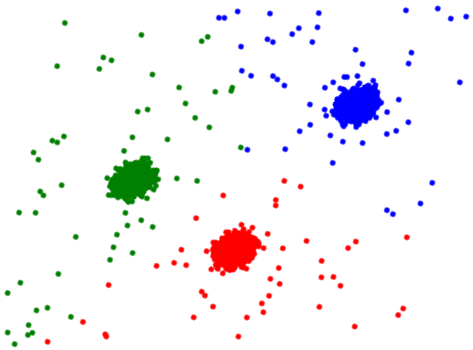}}
	\subfigure[Result with outlier detection (outliers in purple).]{
		\includegraphics[width=0.22\textwidth]{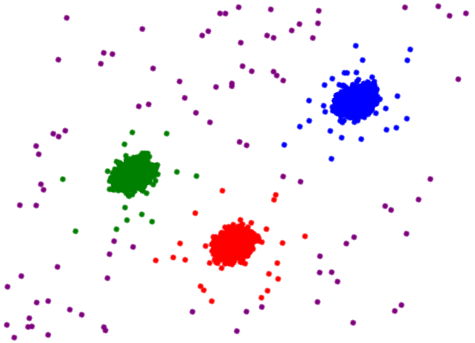}}
	\caption{Clustering results of \method~with and without outlier detection on the second subspace of \textit{syn3o}.}
	\label{fig:outliers}
\end{figure}

This outlier detection procedure is user-friendly since no additional input parameters are necessary. The effectiveness can be seen in Figure \ref{fig:outliers}. The image on the left clearly shows that many points are assigned to a cluster whose center is far away from the actual point. In contrast, our encoding strategy is able to correctly identify most of the outliers (marked in purple), as shown in the right image.

\section{The AutoNR Algorithm}
\label{chap:algorithm}

In this section, we create the example algorithm \method. Therefore, we combine our parameter-free, non-redundant clustering framework with a specific non-redundant clustering algorithm. We exemplary choose \textit{Nr-Kmeans} \cite{mautz2018discovering}, because it uses a relatively simple \textit{k-means}-based objective function. Furthermore, it updates all subspaces simultaneously what supports our idea of full-space clustering executions.

With \textit{k-means}-based clustering, we know that only the distance to the nearest center matters. Here, we can assume that all clusters have an identical sphere of influence, which is the same in all directions. Statistically, this means that the data distribution in each subspace follows a GMM with a tied single-variance covariance matrix. This diagonal matrix has the same variance in each dimension and is valid for all clusters within a subspace, $\Sigma_j = \sigma^2_j I_{m_j}$. Therefore, we can simplify the pdf of the multivariate normal distribution to obtain $L(X_j)$.
\begin{equation*}
\pi_\text{mvnd}(x) =\frac{ \exp \left(-\frac{1}{2}(x-\mu_{j,x})^T\Sigma_j^{-1}(x-\mu_{j,x})\right)}{\sqrt{(2\pi)^{m_j} \det(\Sigma_j)}}
\end{equation*}
\begin{equation*}
\overset{\Sigma_j = \sigma^2_j I_{m_j}}{=} \frac{ \exp \left(-\frac{1}{2\sigma^2_j}||x-\mu_{j,x}||^2_2\right)}{\sqrt{(2\pi \sigma^2_j)^{m_j}}},
\end{equation*}
where $\mu_{j,x}$ is the center to which the object $x$ is assigned to in subspace $j$.
We utilize the maximum log-likelihood estimation to determine the variance $\sigma^2_j$.
\begin{equation*}
\ln \left(\prod_{x \in X_j} \frac{ \exp \left(-\frac{1}{2\sigma^2_j}||x-\mu_{j,x}||^2_2\right)}{\sqrt{(2\pi \sigma^2_j)^{m_j}}} \right)
\end{equation*}
\begin{equation*}
= \sum_{x \in X_j} \ln \left(\frac{ \exp \left(-\frac{1}{2\sigma^2_j}||x-\mu_{j,x}||^2_2\right)}{\sqrt{(2\pi \sigma^2_j)^{m_j}}} \right)
\end{equation*}
\begin{equation}
= - \frac{1}{2} \Big( m_jN \ln (2 \pi \sigma^2_j) + \frac{1}{\sigma^2_j} \sum_{x \in X_j} ||x-\mu_{j,x}||^2_2 \Big)
\label{formula:single_variance}
\end{equation}
By setting the derivative regarding $\sigma^2_j$ to zero we obtain:
\begin{equation}
\sigma^2_j = \frac{1}{m_jN} \sum_{x \in X_j} ||x-\mu_{j,x}||^2_2.
\label{formula:varianz_basic}
\end{equation}
To receive the encoding costs of all objects in a subspace, we sum up the negative binary logarithm of the pdf of each point.

\begin{equation*}
L_\text{mvnd}(X_j) = \sum_{x \in X_j} - \log_2 \left(\frac{ \exp \left(-\frac{1}{2\sigma^2_j}||x-\mu_{j,x}||^2_2\right)}{\sqrt{(2\pi \sigma^2_j)^{m_j}}}\right)
\end{equation*}
\begin{equation*}
= - \frac{1}{\ln (2)} \sum_{x \in X_j} \ln \left(\frac{ \exp \left(-\frac{1}{2\sigma^2_j}||x-\mu_{j,x}||^2_2\right)}{\sqrt{(2\pi \sigma^2_j)^{m_j}}}\right)
\end{equation*}
\begin{equation*}
\overset{(\ref{formula:single_variance})}{=} \frac{1}{2\ln (2)} \Big( m_jN \ln (2 \pi \sigma^2_j) + \frac{1}{\sigma^2_j} \sum_{x \in X_j} ||x-\mu_{j,x}||^2_2 \Big)
\end{equation*}
\begin{equation*}
\overset{(\ref{formula:varianz_basic})}{=} \frac{m_jN}{2\ln (2)} \bigg( 1 + \ln \Big(\frac{2 \pi}{m_jN}\Big) + \ln \Big(\sum_{x \in X_j} ||x-\mu_{j,x}||^2_2\Big)\bigg)
\end{equation*}
The final step is to include $\delta$.
\begin{equation}
L(X_j) = L_\text{mvnd}(X_j) - m_jN\log_2(\delta)
\label{eq:encodingCostsAutoNR}
\end{equation}

Using this encoding, the only distribution-specific parameter is $\sigma^2_j$, thus $p_j=1$. Furthermore, we can use Eq. (\ref{eq:encodingCostsAutoNR}) to calculate the threshold of the distance between a point and its cluster center that determines whether it is an outlier. For clarity, we set $Y = \sum_{x \in X_j} ||x-\mu_{j,x}||^2_2$ and $Z=||o_j-\mu_{j,o_j}||^2_2$.
\begin{equation*}
L_\text{total}(o_j) < L(X_j) - L(X_j \setminus o_j)
\end{equation*}
\begin{equation*}
\begin{gathered}
\overset{(\ref{eq:encodingCostsAutoNR})}{\Rightarrow}
L_\text{total}(o_j) <
\frac{m_jN}{2 \ln (2)} \bigg(1 + \ln \Big(\frac{2 \pi}{m_jN}\Big) + \ln \big(Y\big) \bigg) \\
- \frac{m_j(N-1)}{2 \ln (2)}
\bigg(1 + \ln \Big(\frac{2 \pi}{m_j(N-1)}\Big) + \ln \big(Y-Z\big) \bigg)
\end{gathered}
\end{equation*}
\begin{equation*}
\begin{gathered}
\Rightarrow
\overbrace{\frac{1}{(N-1)}\bigg(\frac{2 \ln (2)L_\text{total}(o_j)}{m_j}- 1 - \ln \Big(\frac{2 \pi}{m_jN}\Big) - \ln \big(Y\big) \bigg)}^{\eqqcolon A} \\
< -\ln \big(N\big) + \ln \big(Y\big) + \ln \big(N-1\big) - \ln \big(Y-Z\big)
\end{gathered}
\end{equation*}
\begin{equation*}
\begin{gathered}
\Rightarrow
Z > Y \big(1 - \frac{N-1}{N}\exp(-A)\big)
\end{gathered}
\end{equation*}
This formulation tells us: If the squared euclidean distance between a point and its center in subspace $j$ is greater than $Y \big(1 - \frac{N-1}{N}\exp(-A)\big)$, then it is an outlier.

Another essential property of our encoding strategy is that it is scaling invariant. Assume that the entire dataset is multiplied by a factor $f$. Then, the precision $\delta$ naturally changes to  $f \delta$. The following applies:
\begin{equation*}
\begin{gathered}
L(f X_j) = L(X_j) + \frac{m_jN}{2 \ln (2)}\ln (f^2) - m_jN\log_2(f) \\
= L(X_j) + 
\frac{m_jN}{\ln (2)}\ln (f) - \frac{m_jN}{\ln (2)}\ln(f)
\end{gathered}
\end{equation*}
Since $\text{max}(\text{dist}_X)$ changes analogously to $\delta$, $L(f \mu_{j, i})=L(\mu_{j, i})$ is also valid. All other costs are trivially independent of factor $f$.

The code, supplement and datasets can be downloaded at \url{www.collinleiber.eu/publications.html}.

\section{Experiments}
\label{chap:experiments}

We conduct experiments on synthetic and real-world datasets. The information regarding the datasets and comparison algorithms can be found in the supplement.
An analysis of the runtime can also be found there.

\begin{table*}[t]
	\centering
	\caption{Experimental results of our method without (\method -) and with (\method +) outlier detection against other algorithms on various datasets. The left side shows the NMI results in \%, the F1 results in \% are shown on the right. All experiments were run ten times, and the average result $\pm$ the standard deviation is stated. The best algorithm for each subspace is marked in bold. The $\dagger$ indicates that that algorithm could not process the dataset due to runtime constraints ($\dagger_R$) or memory issues ($\dagger_M$).}
	\resizebox{\textwidth}{!}{
	\begin{tabular}{l|l|cccccc|cccccc} 
		\toprule
		&&\multicolumn{6}{c|}{NMI (\%)}&\multicolumn{6}{c}{F1 (\%)}\\
		\textbf{Dataset} & \textbf{Subspace} & \method- & \method+ & \textit{NrDipmeans} & \textit{ISAAC} & \textit{MISC} & \textit{FOSSCLU} & \method- & \method+ & \textit{NrDipmeans} & \textit{ISAAC} & \textit{MISC} & \textit{FOSSCLU}\\
		\midrule
		\textbf{syn3} & 1st {\scriptsize ($k_j$=4)} &
		\boldmath{$100\pm0$} & \boldmath{$100\pm0$} & $96\pm7$ & $93\pm15$ & $89\pm19$ & $81\pm1$ & 
		\boldmath{$100\pm0$} & \boldmath{$100\pm0$} & $94\pm10$ & $92\pm18$ & $86\pm24$ & $70\pm2$\\
		
		{\scriptsize (N=5000, d=11)}  & 2nd {\scriptsize ($k_j$=3)} &
		 \boldmath{$100\pm0$} & \boldmath{$100\pm0$} & $87\pm13$ & $95\pm12$ & $92\pm17$ & $65\pm9$ &
		 \boldmath{$100\pm0$} & \boldmath{$100\pm0$} & $85\pm16$ & $94\pm15$ & $89\pm24$ & $60\pm5$\\
		
		 & 3rd {\scriptsize ($k_j$=2)} &
		 \boldmath{$100\pm0$} & \boldmath{$100\pm0$} & $80\pm21$ & \boldmath{$100\pm0$} & $92\pm27$ & $59\pm9$ &
		 \boldmath{$100\pm0$} & \boldmath{$100\pm0$} & $79\pm23$ & \boldmath{$100\pm0$} & $92\pm25$ & $69\pm7$\\
		  
		 \midrule
		 
		 \textbf{syn3o} & 1st {\scriptsize ($k_j$=4)} & 
		 $86\pm10$ & \boldmath{$97\pm0$} & $85\pm10$ & $86\pm19$ & $75\pm18$ & $78\pm2$ &
		 $83\pm18$ & \boldmath{$99\pm0$} & $86\pm12$ & $86\pm22$ & $75\pm22$ & $69\pm1$\\
		 
		 {\scriptsize (N=5150, d=11)} & 2nd {\scriptsize ($k_j$=3)} &
		 $90\pm10$ & \boldmath{$96\pm0$} & $84\pm12$ & $90\pm13$ & $75\pm14$ & $62\pm8$
		 & $91\pm17$ & \boldmath{$99\pm0$} & $85\pm16$ & $89\pm17$ & $75\pm19$ & $59\pm5$\\
		 
		 & 3rd {\scriptsize ($k_j$=2)} &
		 $77\pm20$ & \boldmath{$94\pm0$} & $66\pm17$ & $77\pm31$ & $69\pm30$ & $55\pm6$
		 & $80\pm27$ & \boldmath{$99\pm0$} & $69\pm21$ & $81\pm29$ & $78\pm27$ & $66\pm2$\\
		 
		 \midrule
		 
		 \textbf{Fruits} & Species {\scriptsize ($k_j$=3)} &
		 $85\pm9$ & $83\pm7$ & $82\pm6$ & $21\pm24$ & $56\pm14$ & \boldmath{$90\pm9$}
		 & $89\pm9$ & $87\pm7$ & $87\pm7$ & $56\pm8$ & $59\pm16$ & \boldmath{$93\pm8$}\\
		 
		 {\scriptsize (N=105, d=6)} & Color {\scriptsize ($k_j$=3)} &
		  $17\pm2$ & $18\pm1$ & $17\pm2$ & $6\pm3$ & $13\pm5$ & \boldmath{$19\pm2$}
		 & $47\pm3$ & $44\pm5$ & $47\pm2$ & \boldmath{$49\pm3$} & $42\pm5$ & $40\pm3$\\
		 
		 \midrule
		 
		 \textbf{ALOI} & Shape {\scriptsize ($k_j$=2)} &
		 $62\pm4$ & $64\pm3$ & $65\pm3$ & \boldmath{$70\pm8$} & $38\pm8$ & $58\pm4$
		 & $65\pm2$ & $65\pm1$ & $67\pm1$ & \boldmath{$86\pm6$} & $59\pm5$ & $60\pm4$\\
		 
		 {\scriptsize (N=288, d=611)} & Color {\scriptsize ($k_j$=2)} &
		 $62\pm4$ & $64\pm3$ & \boldmath{$65\pm3$} & $16\pm11$ & $36\pm9$ & $61\pm6$
		 & $65\pm2$ & $65\pm1$ & \boldmath{$66\pm2$} & $55\pm5$ & $58\pm6$ & $62\pm9$\\
		 
		 \midrule
		 
		 \textbf{DSF} & Body-up {\scriptsize ($k_j$=3)} &
		 \boldmath{$100\pm0$} & \boldmath{$100\pm0$} & \boldmath{$100\pm0$} & $71\pm2$ & \boldmath{$100\pm0$} & $73\pm1$
		 & \boldmath{$100\pm0$} & \boldmath{$100\pm0$} & \boldmath{$100\pm0$} & $60\pm3$ & \boldmath{$100\pm0$} & $65\pm2$\\
		 
		 {\scriptsize (N=900, d=400)} & Body-low {\scriptsize ($k_j$=3)} & \boldmath{$100\pm0$} & \boldmath{$100\pm0$} & $71\pm6$ & $71\pm2$ & \boldmath{$100\pm0$} & $42\pm0$
		 & \boldmath{$100\pm0$} & \boldmath{$100\pm0$} & $73\pm5$ & $60\pm3$ & \boldmath{$100\pm0$} & $39\pm1$\\
		 
		 \midrule
		 
		 \textbf{CMUface} & Identity {\scriptsize ($k_j$=20)} &
		 \boldmath{$68\pm4$} & $64\pm4$ & $55\pm0$ & $20\pm4$ & $42\pm10$ & $57\pm7$
		 & \boldmath{$38\pm4$} & $34\pm3$ & $29\pm0$ & $12\pm4$ & $19\pm5$ & $29\pm5$\\
		 
		 {\scriptsize (N=624, d=960)} & Pose {\scriptsize ($k_j$=4)} &
		 \boldmath{$35\pm3$} & $33\pm1$ & $20\pm0$ & $3\pm4$ & $3\pm3$ & $21\pm12$
		 & \boldmath{$45\pm4$} & $42\pm4$ & \boldmath{$45\pm0$} & $34\pm1$ & $30\pm5$ & $41\pm9$\\
		 
		 \midrule
		 
		 \textbf{WebKB} & Category {\scriptsize ($k_j$=4)} &
		 $32\pm2$ & \boldmath{$34\pm3$} & $16\pm4$ & $\dagger_{R}$ & $\dagger_{R}$ & $\dagger_{M}$
		 & $50\pm5$ & \boldmath{$58\pm7$} & $51\pm4$ & $\dagger_{R}$ & $\dagger_{R}$ & $\dagger_{M}$\\
		 
		 {\scriptsize (N=1041, d=323)} & University {\scriptsize ($k_j$=4)} &
		 $56\pm4$ & \boldmath{$57\pm3$} & $24\pm13$ & $\dagger_{R}$ & $\dagger_{R}$ & $\dagger_{M}$
		 & $51\pm2$ & \boldmath{$52\pm3$} & $45\pm5$ & $\dagger_{R}$ & $\dagger_{R}$ & $\dagger_{M}$\\
		 
		 \midrule
		 
		 \textbf{NRLetters} & Letter {\scriptsize ($k_j$=6)} & \boldmath{$100\pm0$} & \boldmath{$100\pm0$} & $95\pm7$ & $87\pm9$ & $92\pm8$ & $82\pm0$
		 & \boldmath{$100\pm0$} & \boldmath{$100\pm0$} & $91\pm12$ & $78\pm13$ & $85\pm11$ & $67\pm2$\\
		 
		 {\scriptsize (N=10000, d=189)} & Color {\scriptsize ($k_j$=3)} &
		 \boldmath{$100\pm0$} & \boldmath{$100\pm0$} & $60\pm30$ & $89\pm18$ & $87\pm18$ & $54\pm18$
		 & \boldmath{$100\pm0$} & \boldmath{$100\pm0$} & $66\pm22$ & $86\pm24$ & $82\pm25$ & $53\pm16$\\
		 
		 & Corner {\scriptsize ($k_j$=4)} &
		 \boldmath{$100\pm0$} & \boldmath{$100\pm0$} & $58\pm28$ & $71\pm11$ & $70\pm11$ & \boldmath{$100\pm0$}
		 & \boldmath{$100\pm0$} & \boldmath{$100\pm0$} & $59\pm23$ & $70\pm11$ & $70\pm11$ & \boldmath{$100\pm0$}\\
		 
		 \midrule
		 
		 \makecell[l]{\textbf{Wine} \\ {\scriptsize (N=178, d=13)}} & Type {\scriptsize ($k_j$=3)} &
		 $76\pm5$ & \boldmath{$85\pm3$} & $34\pm11$ & $0\pm0$ & $32\pm23$ & $84\pm4$
		 & $81\pm6$ & \boldmath{$90\pm4$} & $59\pm5$ & $0\pm0$ & $42\pm29$ & \boldmath{$90\pm4$}\\
		 
		 \midrule
		 	
		 \makecell[l]{\textbf{Optdigits} \\ {\scriptsize (N=5620, d=64)}} & Digit {\scriptsize ($k_j$=10)} &
		 $73\pm1$ & $74\pm1$ & $39\pm14$ & $74\pm7$ & \boldmath{$79\pm1$} & $51\pm2$
		 & $54\pm4$ & $58\pm4$ & $39\pm10$ & $59\pm10$ & \boldmath{$64\pm4$} & $40\pm2$\\
		 
		 \bottomrule
	\end{tabular}
	}
	\label{tab:DataResults}
\end{table*}

\subsection{Quantitative Analyses}

We measure the performance of the algorithms using the ground truth labels of the datasets. In the non-redundant case, the prediction labels ($R^{p} \in \mathbb{N}^{N\times J_p}$) and the ground truth labels ($R^{gt} \in \mathbb{N}^{N \times J_{gt}}$) consist of multiple sets.
Therefore, we report the prediction labels that best match the ground truth of each subspace. For subspace $j$ this equals:
\begin{equation*}
	\text{score}_j(R^{gt}, R^p)=\text{max}(\{\text{metric}(R^{gt}_j, R^p_i) ~ | ~ 1 \le i \le J_p\})
\end{equation*}
We use the \textit{Normalized Mutual Information} (NMI) and the F1 score as metrics. Both produce values between 0 and 1 with an optimum result of 1. Most competitor algorithms are not able to process high-dimensional datasets in an acceptable time ($<$24h). Therefore, we conduct PCA to reduce the number of features while keeping 90\% of the data's variance if $d>50$. Furthermore, we apply standardization (zero mean and unit variance for all features) for \textit{Wine}.

The results of the different experiments are shown in Table \ref{tab:DataResults}. We repeated each experiment ten times and added the average NMI and F1 scores $\pm$ the standard deviation. The NMI and F1 results often differ significantly. This is because the F1 score is more generous if the number of clusters is low.

From these results, we observe that \method~outperforms the other algorithms on almost all datasets. In particular, the comparison methods seem to have problems identifying additional good clusterings after a first high-quality subspace has been found. An example of this situation is \textit{NRLetters}. All algorithms find a first clustering with an average NMI of $>0.89$. However, the quality of the further clusterings decreases more and more for all methods except \method. This shows the value of not initially fixing the subspaces. It is important to note that we are not influencing the order in which the algorithms identify the subspaces. If \method~does not achieve the best results, we are usually in second place (see \textit{Fruits}, \textit{ALOI} and \textit{Optdigits}). 

Outliers negatively affect the clustering results of \method~without outlier detection. This can clearly be seen in the case of \textit{syn3o}. While we achieve perfect results for \textit{syn3}, the quality suffers significantly due to the additional outliers added. Here, our outlier detection significantly improves the results in all three subspaces. This effect also applies to real-world datasets. For example, for \textit{Wine}, the average NMI increases from $0.76$ to $0.85$ and the average F1 score from $0.81$ to $0.90$.

Another positive aspect of \method~with outlier detection is its stability. While the other methods often have a high standard deviation (up to $0.30$), the maximum standard deviation of \method+ across all tested datasets is $0.07$.
Overall, it can be seen that the outlier detection seldom leads to worse results. This means that only rarely are points wrongly identified as outliers. Therefore, we recommend always activating  the outlier detection feature, which is essential for our claim that the procedure is parameter-free.

Based on the results on \textit{Wine} and \textit{Optdigits}, it can be assumed that \method~can also be used for datasets in which no non-redundant clusterings are suspected.

\subsection{Qualitative Analyses}

\begin{figure}[t]
	\centering
	\subfigure['Letter' subspace.]{
		\hspace*{0.05\textwidth}
		\includegraphics[width=0.32\textwidth]{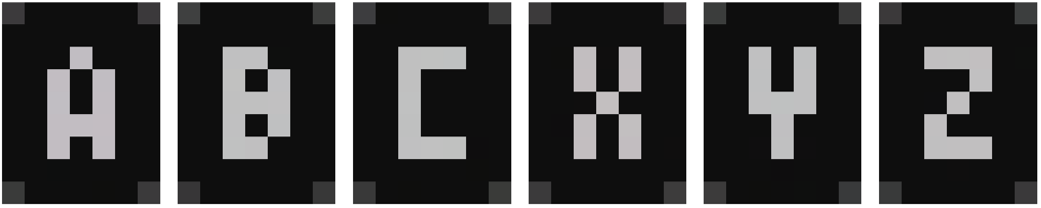}
		\hspace*{0.05\textwidth}
\label{fig:NRLetters_letters}}
	\subfigure['Color' subspace.]{
	\includegraphics[width=0.17\textwidth]{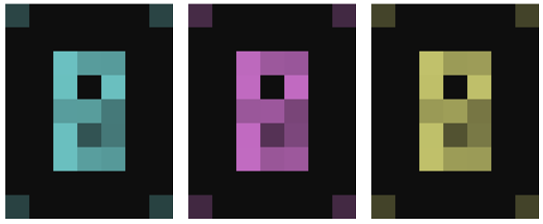}
\label{fig:NRLetters_colors}}
	\subfigure['Corner' subspace.]{
		\includegraphics[width=0.23\textwidth]{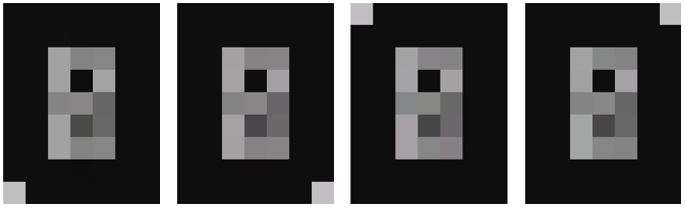}
	\label{fig:NRLetters_corner}}
	\caption{The cluster centers in the three subspaces of \textit{NRLetters} as identified by \method.}
	\label{fig:NRLetters_centers}
\end{figure}

First, we want to mention the high interpretability of the results. For this, we look at the cluster centers found in each subspace of \textit{NRLetters} by \method. These are illustrated in Fig. \ref{fig:NRLetters_centers}.

It can be seen that the first subspace describes the individual letters (\ref{fig:NRLetters_letters}). Here one should note that the colors have been completely extracted from the subspace. The colors are individually described in the second subspace (\ref{fig:NRLetters_colors}). In this case, however, the individual letters are no longer recognizable. The corner markings also receive their individual subspace (\ref{fig:NRLetters_corner}). Here, neither the colors nor the letters are represented. However, only one corner is highlighted, while all corners are slightly tagged in the other subspaces.

Our method shows some problems in clustering \textit{ALOI} correctly. Instead of identifying one subspace each for the shape and color, \method~finds a subspace with the clusters 'red box', 'green box', 'red ball', and 'green ball'. This is also the case with most competitor algorithms. Splitting this space into two would increase the MDL costs. However, if only a single clustering with four clusters is desired, \method~gives very good results.

The opposite is the case with the \textit{Identity} subspace of \textit{CMUface}. Here, splitting the subspace into two spaces reduces the MDL costs. Thus, \method~finds additional substructures in the data that can be used to increase the compression. If we were to merge the two subspaces again, we would get NMI values of over $0.85$ and F1 values of over $0.72$.

\section{Conclusion}
\label{chap:conclusion}

In this work, we introduced a framework to automatically determine the number of subspaces and clusters within subspaces in a non-redundant clustering setting. Additionally, we enhanced the quality of the clustering results by identifying outliers in each of the subspaces. Our framework is easily combinable with different non-redundant clustering approaches.

Experiments show that our exemplary algorithm \method~achieves state-of-the-art results. In particular, the outlier detection feature leads to results outperforming other parameter-free approaches. Since most operations are executed in lower-dimensional subspaces, it is significantly faster than most competitor algorithms for high-dimensional datasets.

At the moment, our framework is only able to handle orthogonal subspaces. Future efforts may attempt to combine our method with multi-view clustering algorithms (e.g. \cite{yao2019multi}). In this setting, dimensions can be used multiple times for several clustering solutions.

\balance

\section*{Acknowledgments}
This work has been partially funded by the German Federal Ministry of Education and Research (BMBF) under Grant No. 01IS18036A. The authors of this work take full responsibility for its content.

The present contribution is supported by the Helmholtz Association under the joint research school 'Munich School for Data Science - MUDS'.

\bibliographystyle{siam}
\bibliography{autonrbib}

\end{document}


\newcommand\relatedversion{}

\title{\Large Supplement - 'Automatic Parameter Selection for Non-Redundant Clustering'\relatedversion}
\author{Collin Leiber\thanks{LMU Munich. \{leiber, mautz, boehm\}@dbs.ifi.lmu.de}
	\and Dominik Mautz$^\ast$
	\and Claudia Plant\thanks{Faculty of Computer Science, ds:UniVie, University of Vienna, Vienna, Austria. claudia.plant@univie.ac.at}
	\and Christian Böhm$^{\ast}$}

\date{}

\maketitle

\fancyfoot[R]{\scriptsize{Copyright \textcopyright\ 2022 by SIAM\\
		Unauthorized reproduction of this article is prohibited}}
\section{Symbols}
\label{sec:symbols}

We use the following symbols in our paper as well as in the supplement:
\begin{table}[h]
		\caption{Description of the used symbols.}
		\centering
		\resizebox{.5\textwidth}{!}{
		\begin{tabular}{|r|r|}
			\hline \makebox{\textbf{Symbol}} & \makebox{\textbf{Interpretation}}\\
			\hline
			$N \in \mathbb{N}$ &  Number of objects\\
			$d \in \mathbb{N}$ &  Dimensionality of the feature space\\
			$J \in \mathbb{N}$ &  Number of subspaces\\
			$V \in \mathbb{R}^{d\times d}$ & Orthogonal (rotation) matrix\\
			$\delta \in \mathbb{R}$ & The precision of the encoding\\
			$I_{m_j} \in \mathbb{N}^{m_j\times m_j}$ & $m_j \times m_j$ identity matrix\\
			$X \subseteq \mathbb{R}^d$ &  Set of all objects\\
			\hline
			$k_j \in \mathbb{N}$ &  Number of clusters in subspace $j$\\
			$m_j \in \mathbb{N}$ &  Dimensionality of subspace $j$\\
			$P_j \in \mathbb{N}^{d\times m_j}$ &  Projection matrix of subspace $j$\\
			$p_j \in \mathbb{N}$ &  \makecell[r]{Number of distribution-specific parameters \\ in subspace $j$}\\
			$X_j \subseteq \mathbb{R}^{m_j}$ & X projected to subspace $j$\\
			$O_j \subseteq X_j$ & Set of outliers in subspace $j$\\
			\hline
			$\mu_{j,i} \in \mathbb{R}^{m_j}$ &  Center of cluster $i$ in subspace $j$\\
			$\Sigma_{j,i}  \in \mathbb{R}^{m_j\times m_j}$ & Covariance matrix of cluster $i$ in subspace $j$\\
			$C_{j,i} \subseteq X_j$ &  Objects of cluster $i$ in subspace $j$\\
			\hline
			$J_p \in \mathbb{N}$ &  Number of predicted subspaces\\
			$J{gt} \in \mathbb{N}$ &  Number of true subspaces\\
			$R^p \in \mathbb{N}^{N \times J_p}$ & Prediction labels matrix\\
			$R^{gt} \in \mathbb{N}^{N \times J_{gt}}$ & Ground truth labels matrix\\
			\hline
		\end{tabular}
	}
	\label{tab:parameters}
\end{table}

\section{Encoding the Constant Values}

At this point, we would like to give a brief intuition on how the constant components of the encoding strategy we present in the paper could be handled.

The number of objects $N$ and dimensionality $d$ can again be encoded using the natural prior for integers, i.e., we need $L^0(N)+L^0(d)$ bits to encode these values. 

In general real values $r$ can be encoded by separately encoding the integer part $\lfloor r \rfloor$ and the decimal places \cite{lee2001introduction}. Here, the precision $\delta$ is required for the decimal places. Hence, the following applies:
\begin{equation*}
L(r) = L^0(\lfloor r \rfloor) - \log_2(\delta).
\end{equation*}

This can be used to encode the $d+1$ values needed to define the hypercube. 
It can also be used to encode $V$. Since all the column vectors of $V$ are orthonormal, they all have a length of $1$. Therefore, all values of a column are less than or equal to $1$ and we can consequently ignore $L^0(\lfloor r \rfloor)$. Moreover, since the orientation is indifferent, the last entry can be calculated using the first $d-1$ entries. Thereby each column loses one degree of freedom. Furthermore, the orthogonal property means that each following column loses an additional degree of freedom. Therefore, we can encode the first column using $-\log_2(\delta)(d-1)$ bits, the second with $-\log_2(\delta)(d-2)$ bits, and so forth. All in all, the code length of $V$ is $- \log_2(\delta)\frac{d(d-1)}{2}$.

From these encodings, it is very easy to see that the values are actually constants that are independent of a particular clustering result. 

\section{Search Space Restrictions}

In this section, we would like to justify our restrictions on the search space with an example. 

In the paper, we say that we restrict the number of clusters in case of a \textit{cluster space} split as follows:
\begin{equation*}
\text{max}(k_{\text{split}_1} , k_{\text{split}_2}) \leq k_{\text{original}} \le k_{\text{split}_1} \cdot k_{\text{split}_2}.
\end{equation*}
Also, we restrict the number of clusters for a \textit{cluster space} merge with the inverted rule.
\begin{equation*}
\text{max}(k_{\text{original}_1} , k_{\text{original}_2}) \leq k_{\text{merge}} \le k_{\text{original}_1} \cdot k_{\text{original}_2}
\end{equation*}

\begin{figure}[t]
	\centering
	\subfigure[Subspace with $4$ clusters.]{
		\includegraphics[width=0.235\textwidth]{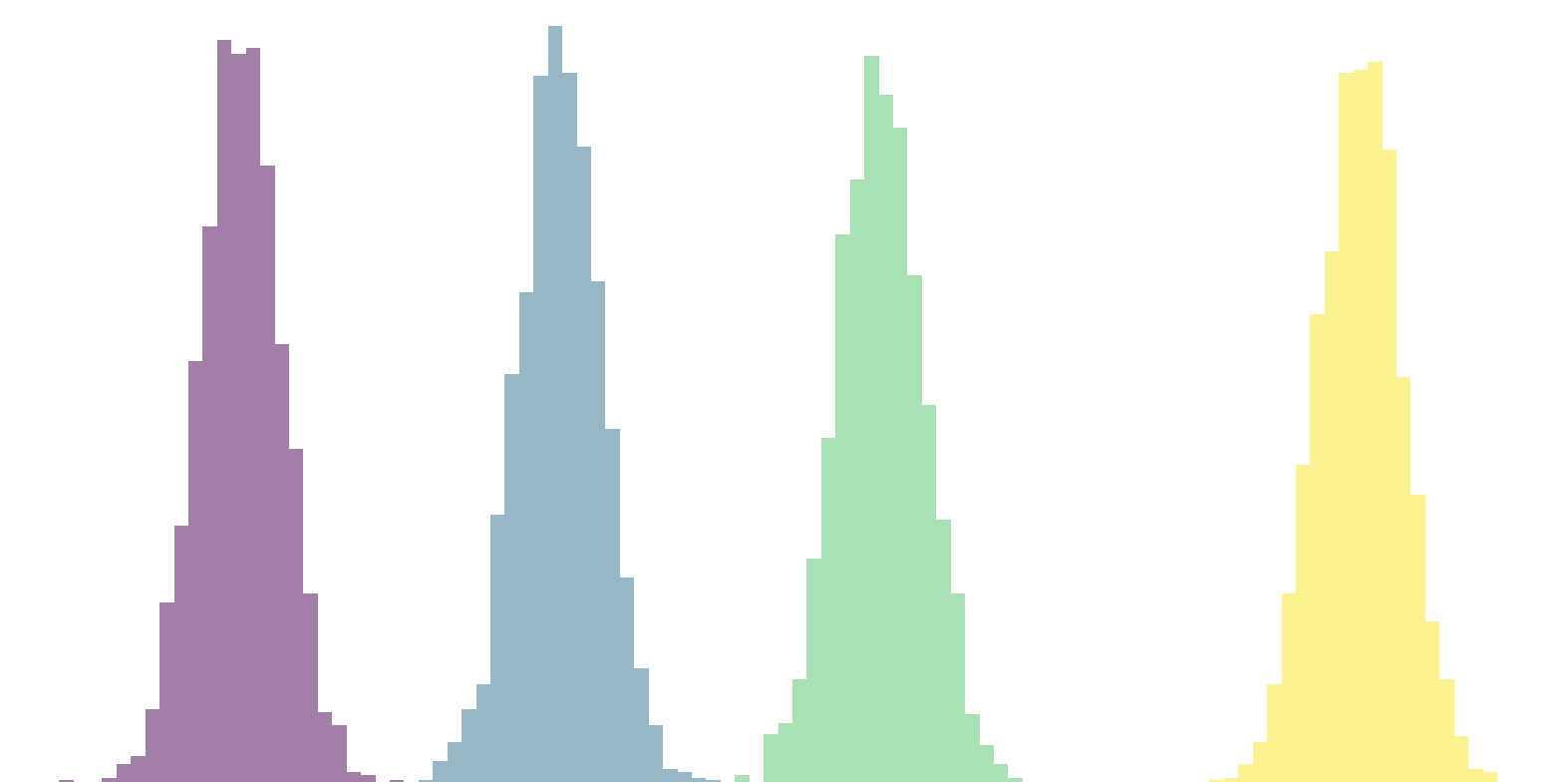}}
	\subfigure[Subspace with $3$ clusters.]{
		\includegraphics[width=0.235\textwidth]{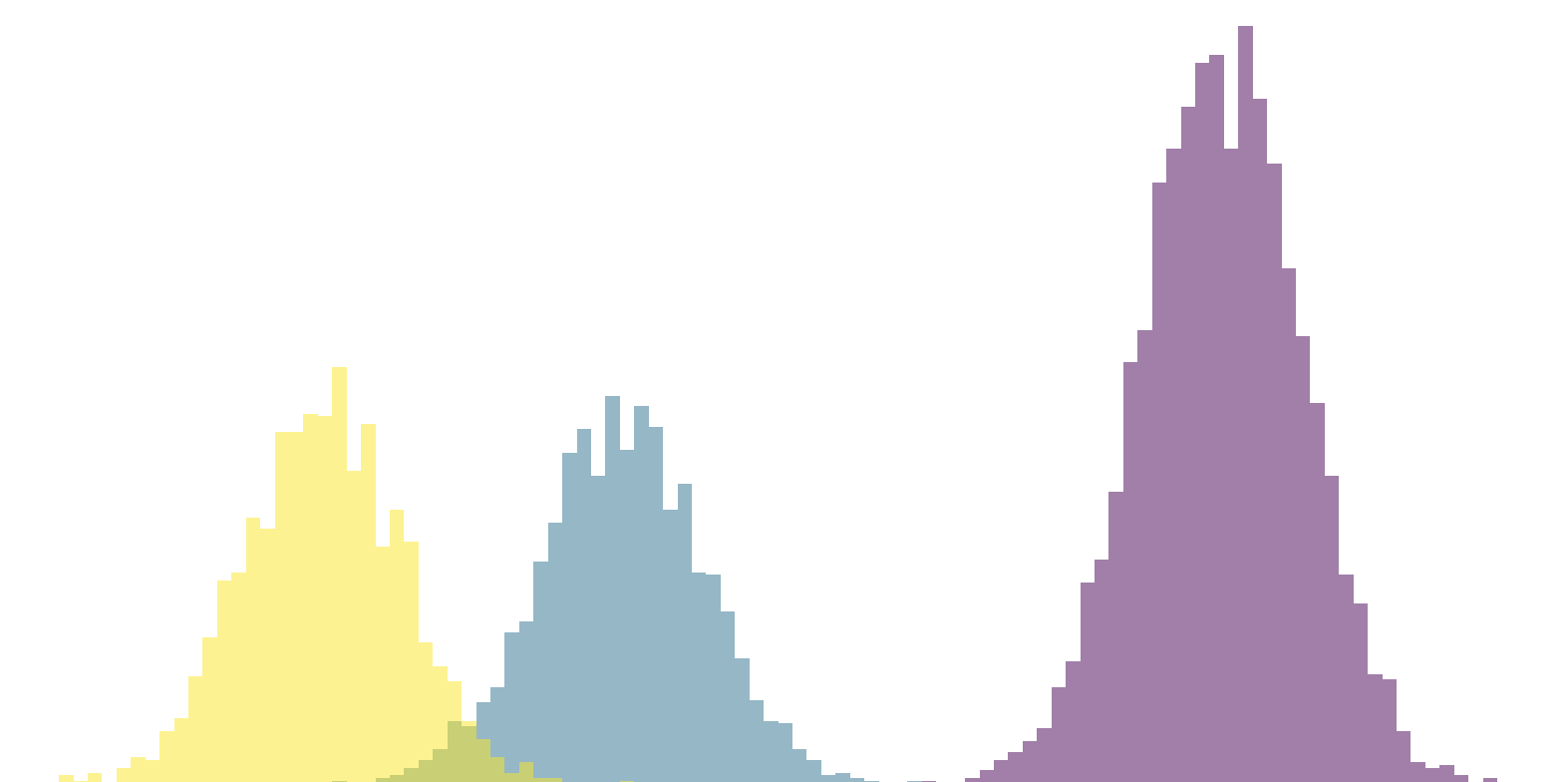}}
	\caption{Histograms of two $1$-dimensional subspaces.}
	\label{fig:1dsubspaces}
\end{figure}

\begin{figure}[t]
	\centering
	\subfigure[Combination with the minimum number of $4$ clusters.]{
		\includegraphics[width=0.235\textwidth]{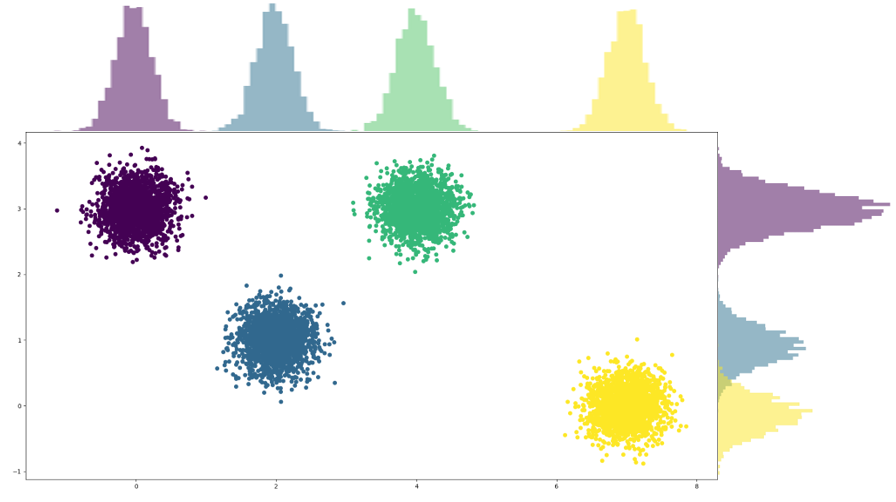}}
	\subfigure[Combination with the maximum number of $12$ clusters.]{
		\includegraphics[width=0.235\textwidth]{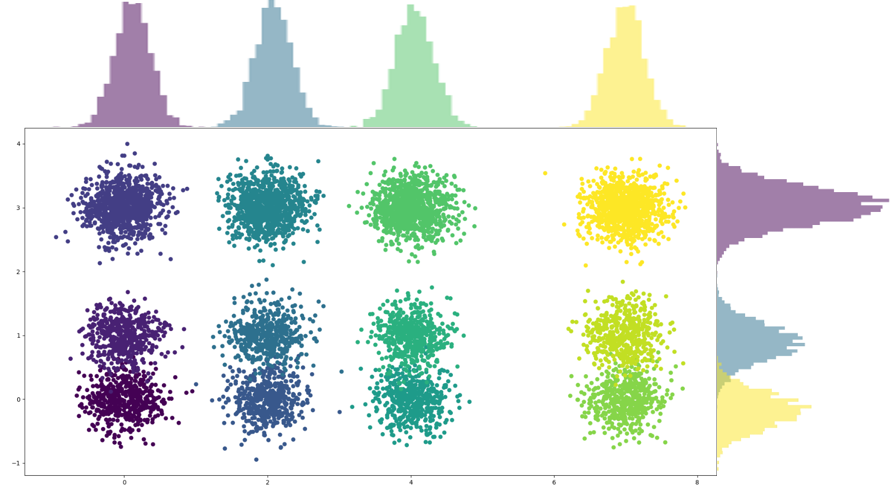}}
	\caption{Two possible combinations of the two subspaces from Figure \ref{fig:1dsubspaces}.}
	\label{fig:2dsubspaces}
\end{figure}

To better understand these rules, assume that we have the $1$-dimensional subspaces shown in Figure \ref{fig:1dsubspaces} with $4$ and $3$ clusters, respectively. If we want to merge these subspaces, we will always get at least 4 clusters, since 4 clusters are already contained in the first subspace. Moreover, there are a maximum of $12$ cluster combinations that can occur. Both extreme situations are illustrated in Figure \ref{fig:2dsubspaces}.

With a \textit{cluster space} split, essentially the same applies, but in reverse. Here, no subspace may be created that already has more than the original number of clusters. Furthermore, $k_{\text{split}_1} \cdot k_{\text{split}_2}$ must be greater than or equal to the original number of clusters. If one of these two rules is not met, subspaces would be created that do not fit the structure of the original subspace.

These rules can also be applied to higher dimensional subspaces.

\section{Pseudo-code}

\begin{algorithm2e}[t]
	\SetAlgoVlined
	\DontPrintSemicolon
	\SetKw{KwGoTo}{go to}
	\label{alg:procedure}
	\SetKwFunction{main}{main}
	\SetKwFunction{merging}{merging}
	\SetKwProg{myalg}{Function}{}{}
	\myalg{\main{dataset $X$}} {
	$V$ = initialize randomly\;
	$R_\text{best}$ = Initial result (single noise space)\;
	// Sort by MDL costs, add noise space last\;
	\textit{sortedSpaces} = sortSubspaces($R_\text{best}$)\; \label{"lnl:sort"}
	\For{\normalfont{$j \in sortedSpaces$}}{
		$X_j=\{xVP_j | x \in X\}$\;
		\uIf{\normalfont{$j$ is cluster space}}{
			// Split space into two cluster spaces\;
			$j_{s_1}, j_{s_2}$ = clusterSpaceSplit($X_j$, $k_j$)\;
		}\ElseIf{\normalfont{$j$ is noise space}}{
		 	// Split space into cluster and noise space\;
			$j_{s_1}, j_{s_2}$ = noiseSpaceSplit($X_j$)\;
		}
		// Check MDL costs\;
		\If{\normalfont{cost($j_{s_1}$) + cost($j_{s_2}$) $<$ cost($j$)}}{
			// Join parameters for full-space execution\;
			$V_\text{tmp}, P_\text{tmp}, \mu_\text{tmp}$ = joinParams($R_\text{best}, V, j_{s_1}, j_{s_2}$)\;
			$R_\text{tmp}$, $V_\text{tmp}$ = fullSpace($X, V_\text{tmp}, P_\text{tmp}, \mu_\text{tmp}$)\;
			// Check full-space MDL costs\;
			\If{\normalfont{cost($R_\text{tmp}$) $<$ cost($R_\text{best}$)}}{
				$R_\text{best}$ $=$ $R_\text{tmp}$; $V = V_\text{tmp}$\;
				\KwGoTo line \ref{"lnl:sort"}\;
			}
		}
	}
	$R_\text{best}, V$ = \merging{$R_\text{best}, V$}\;
	\If{\normalfont{merging was successful}}{
		\KwGoTo line \ref{"lnl:sort"}
	}
	\Return{$R_\text{best}$}
	}
	
	\SetKwProg{myproc}{Function}{}{}
	\myproc{\merging{$R_\text{best}, V$}}{

	\For{\normalfont{each pair ($j_1$, $j_2$) of cluster spaces \label{"lnl:merge"}}}{
		$X_{j_1, j_2}=\{xVP_{j_1, j_2} | x \in X\}$\;
		$j_m$ = clusterSpaceMerge($X_{j_1, j_2}$, $k_{j_1}$, $k_{j_2}$)\;
		// Check MDL costs\;
		\If{\normalfont{cost($j_m$) $<$ cost($j_1$) + cost($j_2$)}}{
			// Join parameters for the full-space execution\;
			$V_\text{tmp}, P_\text{tmp}, \mu_\text{tmp}$ = joinParams($R_\text{best}, V, j_m$)\;
			$R_\text{tmp}, V_\text{tmp}$ = fullSpace($X, V_\text{tmp}, P_\text{tmp}, \mu_\text{tmp}$)\;
			// Check full-space MDL costs\;
			\If{\normalfont{cost($R_\text{tmp}$) $<$ cost($R_\text{best}$)}}{
				$R_\text{best} = R_\text{tmp}$; $V = V_\text{tmp}$\;
			}
		}
	}
	\If{\normalfont{better result found}}{
					\KwGoTo line \ref{"lnl:merge"}
	}
	\Return{$R_\text{best}, V$}			
	}
	\caption{Parameter search algorithm.}
\end{algorithm2e}

In order to determine the number of subspaces and clusters within subspaces for non-redundant clustering, the following steps are executed:
\begin{itemize}
	\item \textit{Noise Space} Split
	\item \textit{Cluster Space} Split
	\item \textit{Cluster Space} Merge
\end{itemize}

Additionally, we regularly combine model parameters to perform a full-space execution. To better understand how all these steps are linked, Algorithm \ref{alg:procedure} can be analyzed.

\section{Implementation Details of AutoNR}
\label{chap:ImplementationDetails}

We want to give additional information regarding the implementation of \method.

Unfortunately, \textit{Nr-Kmeans} introduced another parameter that has to be set by the user. The algorithm optimizes $V$ and $P_j$ through eigenvalue decompositions. Here, the eigenvectors represent the direction, and the signs of the eigenvalues $E$ determine to which subspace the dimensions are assigned. Dimensions not matching the structure of any \textit{cluster space} are assigned to the \textit{noise space}. Consequently, the \textit{noise space} will capture all dimensions corresponding to eigenvalues $\ge 0$. Yet, the supplementary of \cite{mautz2018discovering} states that eigenvectors with a negative eigenvalue close to zero should also be assigned to the \textit{noise space}. An example value is given in the respective publication. However, the optimal threshold changes depending on the input dataset. We want to avoid such hard thresholds in our approach. Therefore, we utilize the described encoding strategy to determine which dimensions should be contained in the cluster and which in the \textit{noise space}.

The rotation matrix $V$ can be updated independently of the new subspace dimensionalities. Therefore, $V$ can be calculated a priori and used in the process to define the new $m_\text{cluster}$ and $m_\text{noise}$. The parameters present in the current iteration of \textit{Nr-Kmeans} can be used to calculate the temporary MDL costs of the model. Since the cluster assignments, cluster centers, and scatter matrices stay constant during this operation, only those costs that depend on the subspace dimensionalities $m_j$ and the projections $P_j$ need to be considered. We start with a \textit{cluster space} that only obtains the dimension corresponding to the lowest eigenvalue and a \textit{noise space} containing the other $|E|-1$ dimensions. The approach is repeated with a rising number of \textit{cluster space} dimensions until the MDL costs exceed the result from the previous iteration or the dimensionality of the \textit{cluster space} equals the number of negative eigenvalues. This means that an initial threshold is no longer necessary

We further utilize the initialization procedure of k-means++ \cite{arthur2007k} to seed the cluster centers.

\section{Evaluation Setup}

\begin{figure*}[t]
	\centering
	\subfigure[Increasing $N$ ($d=6, J=3$).]{
		\includegraphics[width=0.26\textwidth]{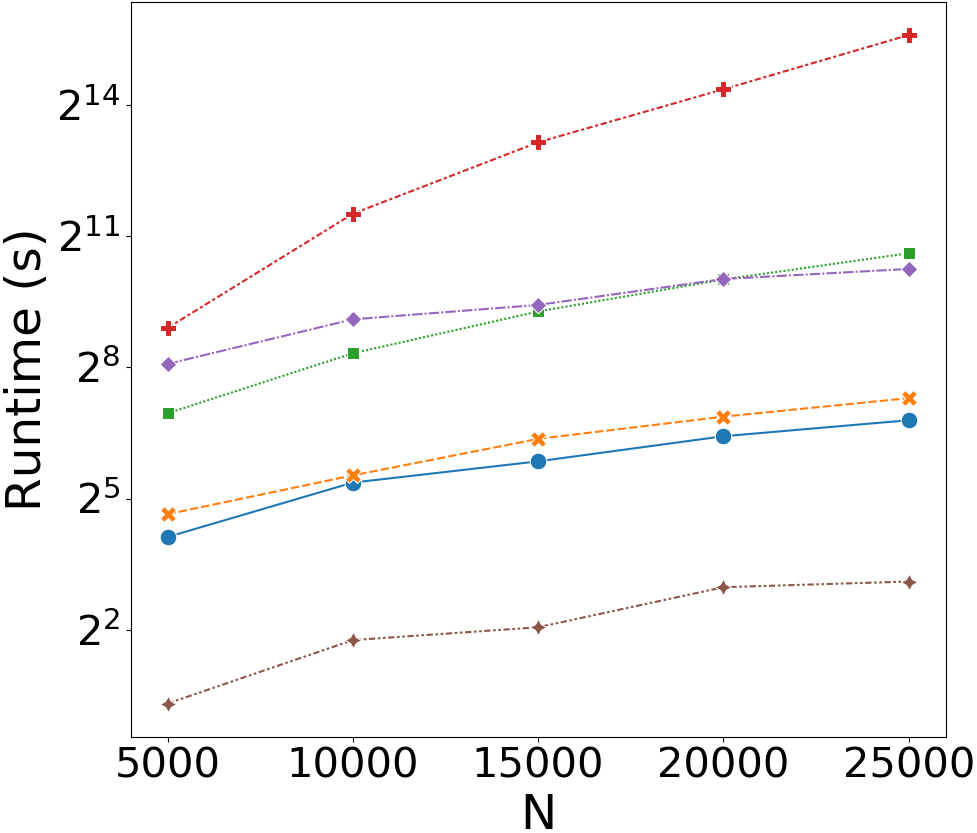} \label{fig:increaseN}}
	\subfigure[Increasing $d$ ($N=5000, J=3$).]{
		\includegraphics[width=0.26\textwidth]{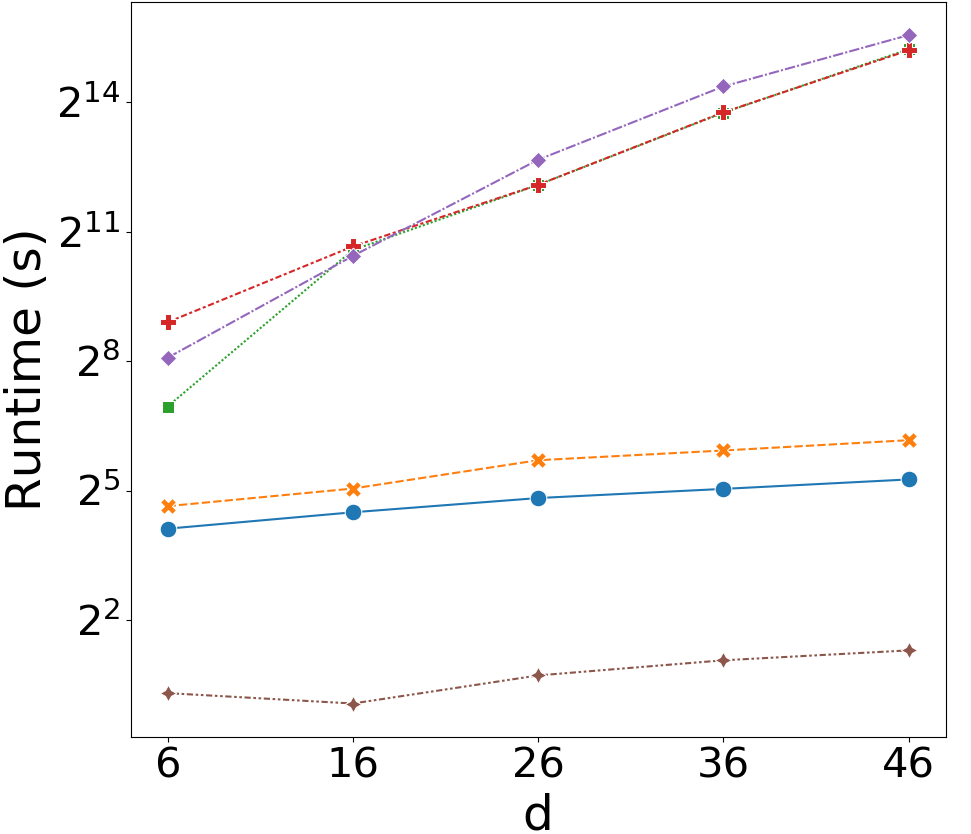} \label{fig:increased}}
	\subfigure[Increasing $J$ ($N=5000, d=2J$).]{
		\includegraphics[width=0.26\textwidth]{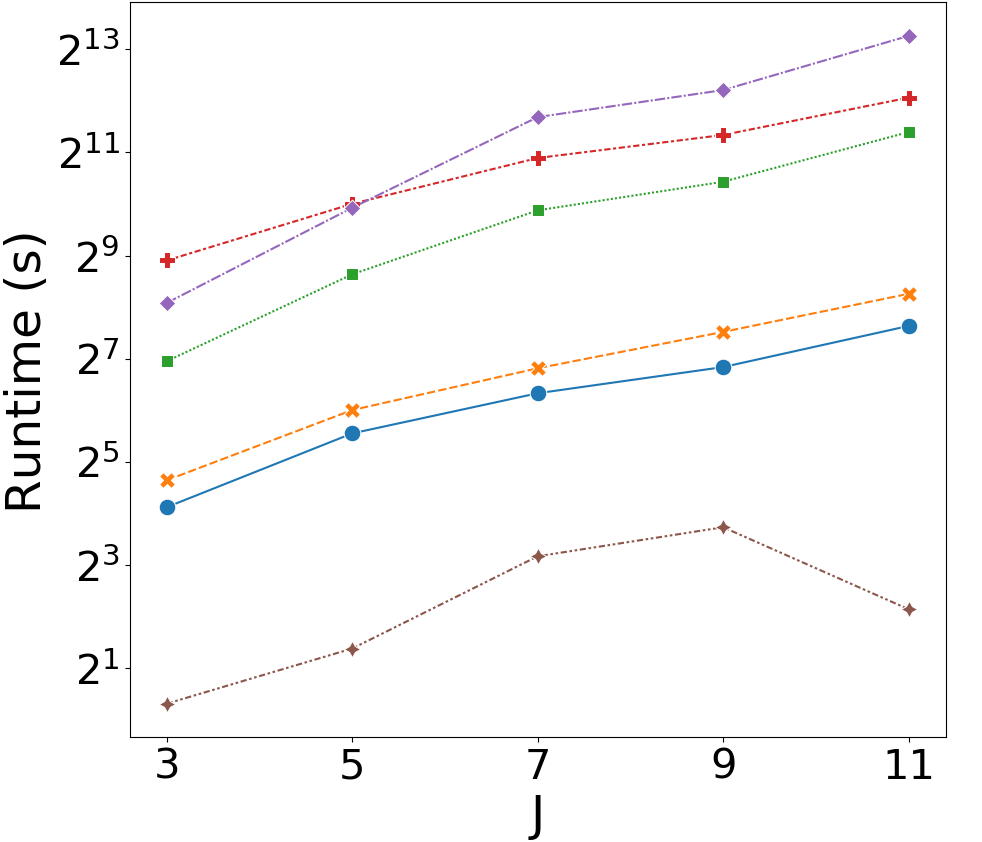} \label{fig:increaseJ}}
	\subfigure[Legend.]{

        \begin{minipage}[t]{0.14\textwidth}
			\centering
			\raisebox{0.3\height}{\includegraphics[width=\textwidth]{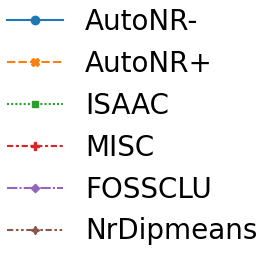}}
	\end{minipage}}
	\caption{Scalability of \method~without (-) and with (+) outlier detection compared to its competitors. All tests are repeated ten times and the mean is stated.}
	\label{fig:runtimeExperiments}
\end{figure*}

\textbf{Datasets:} \textit{syn3} is a synthetic dataset with three subspaces containing $4$, $3$, and $2$ clusters. Each cluster was created using a Gaussian distribution. For \textit{syn3o}, we randomly added 150 uniformly distributed outliers in each subspace.
We additionally created the \textit{NRLetters} dataset. It consists of 10000 $9\times 7$ RGB images of the letters 'A', 'B', 'C', 'X', 'Y', and 'Z' in the colors pink, cyan, and yellow. Moreover, in each image, a corner pixel is highlighted in the color of the letter. This results in three possible clusterings. An extract of this dataset can be seen in the paper in Figure 1.
\textit{Wine} is a real-world dataset from the UCI\footnote{https://archive.ics.uci.edu/ml/index.php} repository with three clusters. The UCI dataset \textit{Optdigits} consists of 5620 $8 \times 8$ images, each representing a digit. The \textit{Fruits} \cite{hu2017finding} dataset was created using 105 images of apples, bananas, and grapes in red, green, and yellow. The images have been preprocessed, resulting in six attributes.
The \textit{Amsterdam Library of Object Image}\footnote{http://aloi.science.uva.nl/} dataset (\textit{ALOI}) contains images of 1000 objects recorded from different angles. For our analysis, we use a common subset of this data consisting of 288 images illustrating the objects 'box' and 'ball' in the colors green and red.
\textit{Dancing Stick Figures} \cite{gunnemann2014smvc} (\textit{DSF}) is a dataset containing 900 $20\times 20$ images. It comprises two subspaces describing three upper- and three lower-body motions.
\textit{CMUface} is again taken from the UCI repository and is composed of 640 $30\times 32$ gray-scaled images showing 20 persons in four different poses (up, straight, left, right). 
Among those images, 16 show glitches resulting in 624 useful objects. 
The \textit{WebKB}\footnote{http://www.cs.cmu.edu/~webkb/} dataset contains 1041 Html documents from four universities. These web pages belong to one of four categories. We preprocessed the data using stemming and removed stop words and words with a document frequency $<1\%$. Afterward, we removed words with a variance $<0.25$, resulting in 323 features.

\textbf{Comparison Methods:} We compare the results of \method~without (\method-) and with (\method+) outlier detection against the parameter-free algorithms \textit{ISAAC} \cite{ye2016generalized} and \textit{MISC} \cite{wang2019multiple} as well as \textit{NrDipmeans} \cite{mautz2020non}. Furthermore, we extend the subspace clustering approach \textit{FOSSCLU} \cite{goebl2014finding} to iteratively identify new subspaces by removing the subspaces found in previous iterations. For \textit{NrDipmeans} and \textit{FOSSCLU} we have to state the desired number of subspaces. In case of \textit{FOSSCLU} we need to define limits for $m_j$ and $k_j$. We set those to $1\le m_j \le 3$ and $2\le k_j \le 10$. We wanted to set the upper bound of $k_j$ to $20$ for \textit{CMUface}, so \textit{FOSSCLU} would be able to determine all parameters correctly. Unfortunately, this leads to memory issues. Where required, \method~runs $15$ executions of \textit{Nr-Kmeans}. The significance for \textit{NrDipmeans} is set to $0.01$.

Experiments are conducted using the Scala implementations of \textit{Nr-Kmeans} and \textit{NrDipmeans} and the Matlab implementations of \textit{ISAAC} and \textit{MISC} as referenced in \cite{mautz2018discovering}, \cite{mautz2020non}, \cite{ye2016generalized} and \cite{wang2019multiple} respectively. Regarding \textit{FOSSCLU}, we extend the Java version referenced in \cite{goebl2014finding} as described above. \method~is implemented in Python.

\section{Runtime Analysis}
We conduct runtime experiments on datasets with a rising number of objects $N$, dimensions $d$, and subspaces $J$. The created \textit{cluster spaces} are always two-dimensional and contain three Gaussian clusters each. 

All experiments are performed on a computer with an Intel Core i7-8700 3.2 GHz processor and 32GB RAM. The runtime results again correspond to the average of ten consecutive executions. The outcomes are illustrated in Figure \ref{fig:runtimeExperiments}.

The charts show that our approach is well applicable to high-dimensional datasets. The runtime increases only slightly with additional \textit{noise space} dimensions (\ref{fig:increased}). \textit{ISAAC} and \textit{MISC} have to conduct an ISA which does not scale well to high-dimensional datasets. \textit{FOSSCLU} has to perform Givens rotations multiple times, which is an expensive operation. On the other hand, our framework performs most steps in lower-dimensional subspaces where the overall dimensionality has no significant influence. If additional \textit{cluster spaces} accompany a higher dimensionality, the runtimes of all algorithms behave similarly (\ref{fig:increaseJ}). For large datasets, the differences in runtime are also much less prominent (\ref{fig:increaseN}). Only \textit{MISC} needs significantly more time because a kernel graph regularized semi-nonnegative matrix factorization has to be performed.

Due to the additional operations required to calculate the outlier distance threshold for each subspace in each iteration, the execution of \method~with outlier detection expectably takes more time than without. Furthermore, the cluster centers and covariance matrices are updated after each outlier detection procedure.

\textit{NrDipmeans} is the fastest in all experiments. However, it must be noted that \textit{NrDipmeans} knows the correct number of subspaces and therefore does not need to run tests to identify $J$. Furthermore, in the case of $J=11$, it settles with the initial two clusters in each subspace and does not invest time in finding better structures. Therefore, it seems to have problems running with a high $J$. \method, on the other hand, almost always correctly identifies all clusters in all subspaces.

Our procedure could be further accelerated by, for example, parallelizing the multiple executions of \textit{Nr-Kmeans} with identical parameters.

\section{Comparison to \textit{Nr-Kmeans}}

We perform additional experiments using the original \textit{Nr-Kmeans} algorithm, to show that the good experimental results are based on our proposal and not merely on the integration of \textit{Nr-Kmeans}. The new results are shown in Table \ref{tab:DataResultsNrKmeans}. As in the paper, we repeated each experiment ten times and added the average score $\pm$ the standard deviation to the table.

\begin{table*}[t]
	\centering
	\caption{Results of \method - and \method + compared to our \textit{Nr-Kmeans} version, where the dimensionality of the \textit{noise space} is determined through MDL (see supplement Section \ref{chap:ImplementationDetails}), and the original \textit{Nr-Kmeans} implementation on various datasets. The left side shows the NMI results in \%, the F1 results in \% are shown on the right. All experiments were run ten times, and the average result $\pm$ the standard deviation is stated. The best algorithm for each subspace is marked in bold.}
	\resizebox{0.72\textwidth}{!}{
		\begin{tabular}{l|l|cccc|cccc} 
			\toprule
			&&\multicolumn{4}{c|}{NMI (\%)} & \multicolumn{4}{c}{F1 (\%)}\\
			\textbf{Dataset} & \textbf{Subspace} & \method- & \method+ & \makecell[c]{\textit{Nr-Kmeans}\\(MDL-based \textit{noise space})} & \makecell[c]{\textit{Nr-Kmeans}\\(Original)}
			& \method- & \method+ & \makecell[c]{\textit{Nr-Kmeans}\\(MDL-based \textit{noise space})} & \makecell[c]{\textit{Nr-Kmeans}\\(Original)}\\
			\midrule
			\textbf{syn3} & 1st {\scriptsize ($k_j$=4)} &
			\boldmath{$100\pm0$} & \boldmath{$100\pm0$} & $73\pm20$ & $59\pm16$ &
			\boldmath{$100\pm0$} & \boldmath{$100\pm0$} & $72\pm18$ & $61\pm11$ \\
			
			{\scriptsize (N=5000, d=11)}  & 2nd {\scriptsize ($k_j$=3)} &
			\boldmath{$100\pm0$} & \boldmath{$100\pm0$} & $81\pm13$ & $75\pm17$ &
			\boldmath{$100\pm0$} & \boldmath{$100\pm0$} & $82\pm12$ & $78\pm16$ \\
			
			& 3rd {\scriptsize ($k_j$=2)} &
			\boldmath{$100\pm0$} & \boldmath{$100\pm0$} & $72\pm22$ & $75\pm17$ &
			\boldmath{$100\pm0$} & \boldmath{$100\pm0$} & $79\pm16$ & $81\pm14$ \\
			
			\midrule
			
			\textbf{syn3o} & 1st {\scriptsize ($k_j$=4)} & 
			$86\pm10$ & \boldmath{$97\pm0$} & $62\pm17$ & $54\pm13$ &
			$83\pm18$ & \boldmath{$99\pm0$} & $64\pm14$ & $57\pm10$ \\
			
			{\scriptsize (N=5150, d=11)} & 2nd {\scriptsize ($k_j$=3)} &
			$90\pm10$ & \boldmath{$96\pm0$} & $67\pm15$ & $69\pm11$ &
			$91\pm17$ & \boldmath{$99\pm0$} & $73\pm14$ & $75\pm9$ \\
			
			& 3rd {\scriptsize ($k_j$=2)} &
			$77\pm20$ & \boldmath{$94\pm0$} & $56\pm10$ & $68\pm13$ &
			$80\pm27$ & \boldmath{$99\pm0$} & $63\pm10$ & $77\pm13$ \\
			
			\midrule
			
			\textbf{Fruits} & Species {\scriptsize ($k_j$=3)} &
			\boldmath{$85\pm9$} & $83\pm7$ & $70\pm14$ & $74\pm11$ &
			\boldmath{$89\pm9$} & $87\pm7$ & $78\pm11$ & $78\pm11$ \\
			
			{\scriptsize (N=105, d=6)} & Color {\scriptsize ($k_j$=3)} &
			$17\pm2$ & \boldmath{$18\pm1$} & $15\pm2$ & $16\pm2$ &
			\boldmath{$47\pm3$} & $44\pm5$ & $42\pm3$ & $44\pm3$ \\
			
			\midrule
			
			\textbf{ALOI} & Shape {\scriptsize ($k_j$=2)} &
			$62\pm4$ & \boldmath{$64\pm3$} & $47\pm26$ & $54\pm30$ &
			$65\pm2$ & $65\pm1$ & $73\pm13$ & \boldmath{$77\pm15$} \\
			
			{\scriptsize (N=288, d=611)} & Color {\scriptsize ($k_j$=2)} &
			$62\pm4$ & \boldmath{$64\pm3$} & $34\pm0$ & $31\pm10$ &
			$65\pm2$ & $65\pm1$ & \boldmath{$66\pm0$} & \boldmath{$66\pm0$} \\
			
			\midrule
			
			\textbf{DSF} & Body-up {\scriptsize ($k_j$=3)} &
			\boldmath{$100\pm0$} & \boldmath{$100\pm0$} & $70\pm20$ & $81\pm26$ &
			\boldmath{$100\pm0$} & \boldmath{$100\pm0$} & $77\pm16$ & $85\pm20$ \\
			
			{\scriptsize (N=900, d=400)} & Body-low {\scriptsize ($k_j$=3)} &
			\boldmath{$100\pm0$} & \boldmath{$100\pm0$} & $63\pm22$ & $56\pm30$ &
			\boldmath{$100\pm0$} & \boldmath{$100\pm0$} & $70\pm17$ & $68\pm22$ \\
			
			\midrule
			
			\textbf{CMUface} & Identity {\scriptsize ($k_j$=20)} &
			$68\pm4$ & $64\pm4$ & \boldmath{$78\pm6$} & $75\pm7$ &
			$38\pm4$ & $34\pm3$ & \boldmath{$57\pm9$} & $52\pm9$ \\
			
			{\scriptsize (N=624, d=960)} & Pose {\scriptsize ($k_j$=4)} &
			\boldmath{$35\pm3$} & $33\pm1$ & $28\pm8$ & $26\pm6$ &
			\boldmath{$45\pm4$} & $42\pm4$ & $41\pm10$ & $37\pm10$ \\
			
			\midrule
			
			\textbf{WebKB} & Category {\scriptsize ($k_j$=4)} &
			$32\pm2$ & \boldmath{$34\pm3$} & $32\pm3$ & $30\pm2$ &
			$50\pm5$ & \boldmath{$58\pm7$} & $48\pm3$ & $48\pm2$ \\
			
			{\scriptsize (N=1041, d=323)} & University {\scriptsize ($k_j$=4)} &
			$56\pm4$ & \boldmath{$57\pm3$} & $47\pm8$ & $45\pm7$ &
			$51\pm2$ & $52\pm3$ & \boldmath{$54\pm7$} & $52\pm3$ \\
			
			\midrule
			
			\textbf{NRLetters} & Letter {\scriptsize ($k_j$=6)} & 
			\boldmath{$100\pm0$} & \boldmath{$100\pm0$} & $85\pm9$ & $83\pm9$ &
			\boldmath{$100\pm0$} & \boldmath{$100\pm0$} & $78\pm13$ & $72\pm12$ \\
			
			{\scriptsize (N=10000, d=189)} & Color {\scriptsize ($k_j$=3)} &
			\boldmath{$100\pm0$} & \boldmath{$100\pm0$} & $52\pm29$ & $39\pm25$ &
			\boldmath{$100\pm0$} & \boldmath{$100\pm0$} & $61\pm22$ & $52\pm18$ \\
			
			& Corner {\scriptsize ($k_j$=4)} &
			\boldmath{$100\pm0$} & \boldmath{$100\pm0$} & $57\pm26$ & $48\pm25$ &
			\boldmath{$100\pm0$} & \boldmath{$100\pm0$} & $61\pm23$ & $51\pm19$ \\
			
			\midrule
			
			\makecell[l]{\textbf{Wine} \\ {\scriptsize (N=178, d=13)}} & Type {\scriptsize ($k_j$=3)} &
			$76\pm5$ & $85\pm3$ & \boldmath{$87\pm2$} & $79\pm15$ &
			$81\pm6$ & $90\pm4$ & \boldmath{$92\pm2$} & $87\pm10$ \\
			
			\midrule
			
			\makecell[l]{\textbf{Optdigits} \\ {\scriptsize (N=5620, d=64)}} & Digit {\scriptsize ($k_j$=10)} &
			$73\pm1$ & \boldmath{$74\pm1$} & $72\pm2$ & $70\pm2$ &
			$54\pm4$ & $58\pm4$ & \boldmath{$66\pm3$} & $64\pm3$ \\
			
			\bottomrule
		\end{tabular}
	}
	\label{tab:DataResultsNrKmeans}
\end{table*}

\method~returns superior results in most experiments, even though \textit{Nr-Kmeans} already knows the correct number of subspaces and clusters for each subspace. Only for the non-redundant dataset \textit{ALOI} does the original \textit{Nr-Kmeans} perform better regarding the F1 score. This case, however, has already been mentioned in the paper. The biggest advantage of our application is the fact that it discovers structures one by one while preserving the ability to adjust already found subspaces. This gives great flexibility, so that possible errors can be compensated in a following iteration. Another advantage is the definition of the \textit{noise space} using MDL (see supplement Section \ref{chap:ImplementationDetails}), as it can be seen with the datasets \textit{CMUface}, \textit{WebKB}, \textit{NRLetters}, \textit{Wine} and \textit{Optdigits}.

One could argue that the multiple repetitions of \textit{Nr-Kmeans} included in each run of \method~strongly favor our algorithm. However, we would like to counter this by stating that \textit{Nr-Kmeans} by itself is often unable to achieve a perfect result just once (e.g., for \textit{syn3}).  In contrast, \method~often assigns the points to the correct clusters every time. This shows that the iterative identification of subspaces can be beneficial, with the effect becoming stronger the more subspaces there are.

\balance
\section{Importance of Correct Parametrization}

\begin{figure}[t]
	\centering
	\subfigure{
		\includegraphics[width=0.235\textwidth]{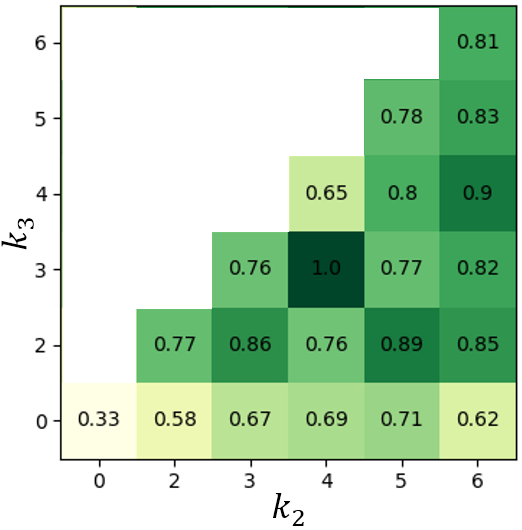}}
	\subfigure{
		\includegraphics[width=0.235\textwidth]{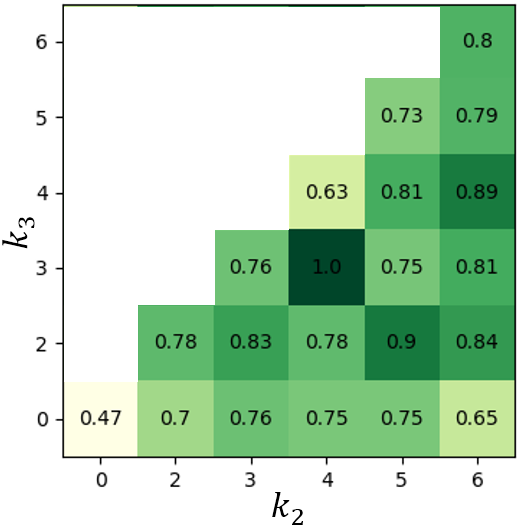}}
	\caption{Results of various parametrizations of \textit{Nr-Kmeans} on \textit{NRLetters}. $k_1$ is set to $6$. The left image shows the NMI, and the right the F1 results.}
	\label{fig:parameterImportance}
\end{figure}

To better assess the importance of a correct parametrization of non-redundant clustering approaches, we perform another  experiment. Suppose that we know that \textit{NRLetters} comprises six different letters. We know nothing about the other clustering possibilities. Therefore, we try different parameters for \textit{Nr-Kmeans} using a brute-force search. Here, we assume that no subspace contains more than six clusters. The NMI and F1 results can be seen in Figure \ref{fig:parameterImportance}. To arrive at a single number that indicates the quality of a non-redundant clustering result, we compute the average result over all three subspaces.
\begin{equation*}
	\text{score}(R^{gt}, R^p) = \frac{1}{J_{gt}} \sum_{1\le j \le J_{gt}} \textit{score}_j(R^{gt}, R^p),
\end{equation*}
where $\textit{score}_j(R^{gt}, R^p)$ is the evaluation method as described in the paper and $J_{gt}$ is the number of label sets in the ground truth. Each run is repeated ten times and the best result is added to the heatmap.

We see that the quality of the results deteriorates away from the optimum ($k_2=4, k_3=3$) even though we know the correct number of clusters in the first subspace. This shows the value of our framework, which achieved a perfect result in all ten iterations and that, without prior knowledge.

\bibliographystyle{siam}
\bibliography{autonrbib}